 \documentclass[final,5p,times,twocolumn]{elsarticle}

\usepackage{amssymb}
\usepackage{caption}
\usepackage{subcaption}
\usepackage{graphicx}
\usepackage{booktabs}
\usepackage{multirow}
\usepackage{amsmath}
\usepackage{xcolor}
\usepackage{comment}
\DeclareMathOperator*{\argmax}{argmax} % thin space, limits underneath in displays

\journal{Future Generation Computer Systems}

\begin{document}

\begin{frontmatter}

%% Title, authors and addresses

%% use the tnoteref command within \title for footnotes;
%% use the tnotetext command for theassociated footnote;
%% use the fnref command within \author or \address for footnotes;
%% use the fntext command for theassociated footnote;
%% use the corref command within \author for corresponding author footnotes;
%% use the cortext command for theassociated footnote;
%% use the ead command for the email address,
%% and the form \ead[url] for the home page:
%% \title{Title\tnoteref{label1}}
%% \tnotetext[label1]{}
%% \author{Name\corref{cor1}\fnref{label2}}
%% \ead{email address}
%% \ead[url]{home page}
%% \fntext[label2]{}
%% \cortext[cor1]{}
%% \affiliation{organization={},
%%             addressline={},
%%             city={},
%%             postcode={},
%%             state={},
%%             country={}}
%% \fntext[label3]{}

\title{Reducing Inference Energy Consumption Using Dual Complementary CNNs}

%% use optional labels to link authors explicitly to addresses:
%% \author[label1,label2]{}
%% \affiliation[label1]{organization={},
%%             addressline={},
%%             city={},
%%             postcode={},
%%             state={},
%%             country={}}
%%
%% \affiliation[label2]{organization={},
%%             addressline={},
%%             city={},
%%             postcode={},
%%             state={},
%%             country={}}

\author[inst1]{Michail Kinnas}

\affiliation[inst1]{organization={Dept. Informatics \& Telematics, Harokopio University of Athens},%Department and Organization
            addressline={Omirou 9}, 
            city={Tavros},
            postcode={177 78}, 
            %state={State One},
            country={Greece}}

\author[inst2]{John Violos}
\author[inst2]{Ioannis Kompatsiaris}
\author[inst2]{Symeon Papadopoulos}
\affiliation[inst2]{organization={Information Technologies Institute, Centre for Research and Technology},
            addressline={6th km Harilaou - Thermi}, 
            city={Thessaloniki},
            postcode={57001}, 
            country={Greece}}

\begin{abstract}
% Energy efficiency of Convolutional Neural Networks (CNNs) has become an important area of research, with various strategies being developed to minimize the power consumption of these models. Previous efforts, including techniques like model pruning, quantization, and hardware optimization, have made significant strides in this direction. However, there remains a need for more effective on device AI solutions that balance energy efficiency with model performance. In this paper, we propose a novel approach to reduce the energy requirements of inference of CNNs. Our methodology employs two small Complementary CNNs that collaborate with each other by covering each other's "weaknesses" in predictions. If the confidence for a prediction of the first CNN is considered low, the second CNN is invoked with the aim of producing a higher confidence prediction. This dual-CNN setup significantly reduces energy consumption compared to using a single large deep CNN. Additionally, we propose a memory component that retains previous classifications for identical inputs, bypassing the need to re-invoke the CNNs for the same input, further saving energy. Our experiments on a Jetson Nano computer demonstrate an energy reduction of up to 85.8\% for CIFAR-10 samples and up to 80.9\% for ImageNet, achieved on modified datasets where each sample was duplicated once. These findings indicate that leveraging a complementary CNN pair along with a memory component effectively reduces inference energy while maintaining high accuracy.

% Major review
Energy efficiency of Convolutional Neural Networks (CNNs) has become an important area of research, with various strategies being developed to minimize the power consumption of these models. Previous efforts, including techniques like model pruning, quantization, and hardware optimization, have made significant strides in this direction. However, there remains a need for more effective on device AI solutions that balance energy efficiency with model performance. In this paper, we propose a novel approach to reduce the energy requirements of inference of CNNs. Our methodology employs two small Complementary CNNs that collaborate with each other by covering each other's "weaknesses" in predictions. If the confidence for a prediction of the first CNN is considered low, the second CNN is invoked with the aim of producing a higher confidence prediction. This dual-CNN setup significantly reduces energy consumption compared to using a single large deep CNN. Additionally, we propose a memory component that retains previous classifications for identical inputs, bypassing the need to re-invoke the CNNs for the same input, further saving energy. Our experiments on a Jetson Nano computer demonstrate an energy reduction of up to 85.8\% achieved on modified datasets where each sample was duplicated once. These findings indicate that leveraging a complementary CNN pair along with a memory component effectively reduces inference energy while maintaining high accuracy.
\end{abstract}

%%Graphical abstract
\begin{graphicalabstract}
\includegraphics[width=\textwidth]{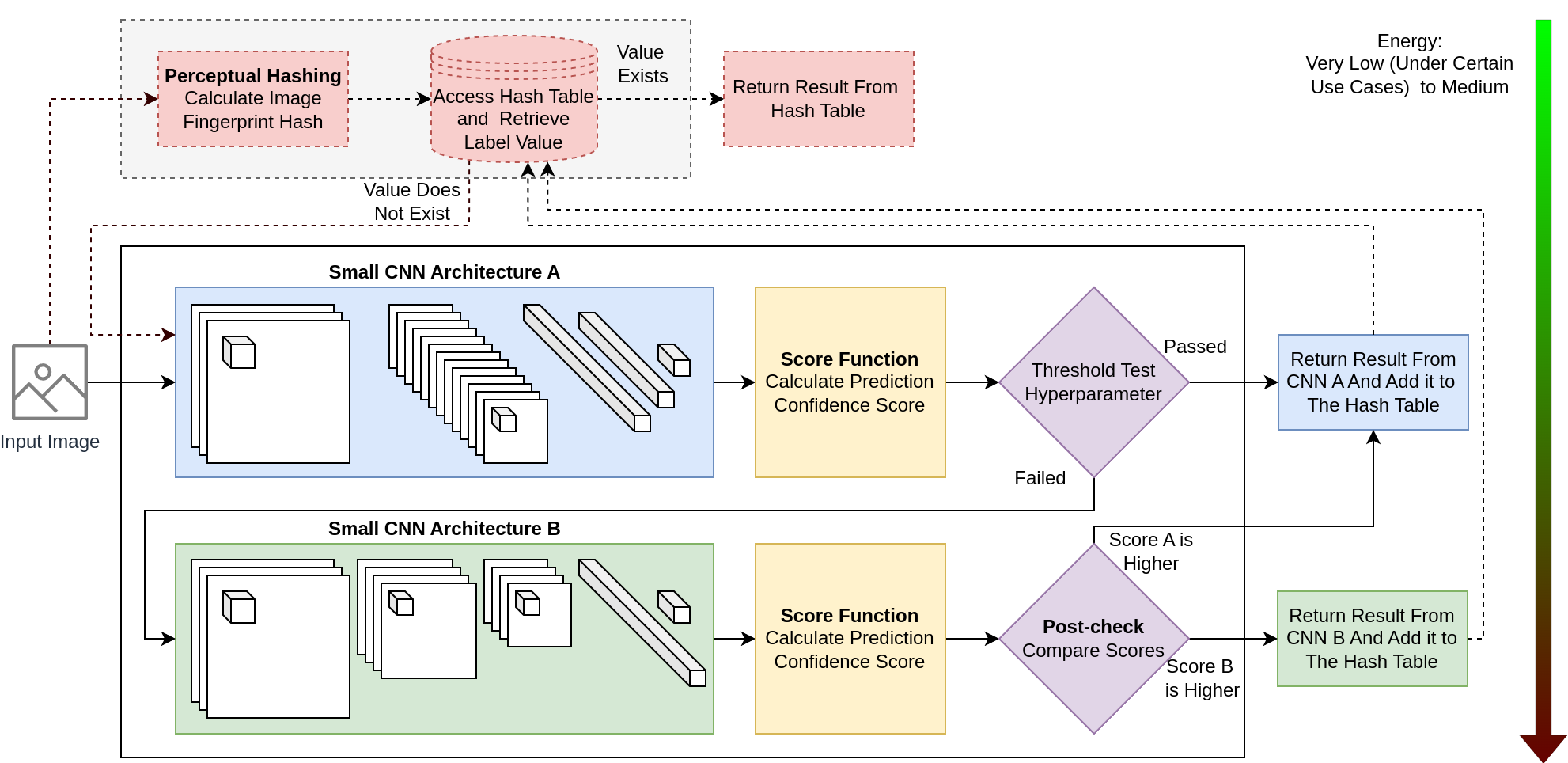}
\end{graphicalabstract}

%%Research highlights
\begin{highlights}
\item Introduced the concept of complementarity in Neural Networks.
\item Proposed an on-device AI methodology based on the synergy of two small, complementary CNNs. This method is optimized for Edge devices with limited computational resources.
\item For the collaboration of two small CNNs a confidence-based mechanism used. With this mechanism the system dynamically selects the most accurate predictions from the paired CNNs, enhancing the efficiency of on-device inference.
\item Conducted experimental evaluations using four datasets on a Jetson Nano device, with power consumption monitored via a power meter.
\item Demonstrated that the collaboration between the two smaller CNNs significantly reduces energy consumption and response time, while maintaining accuracy levels comparable to larger, state-of-the-art CNN models.
\end{highlights}

\begin{keyword}
%% keywords here, in the form: keyword \sep keyword
On-device AI Applications \sep Convolutional Neural Networks \sep Energy Consumption \sep Complementarity \sep Confidence Score \sep Perceptual Hash 
%% PACS codes here, in the form: \PACS code \sep code
%\PACS 0000 \sep 1111
%% MSC codes here, in the form: \MSC code \sep code
%% or \MSC[2008] code \sep code (2000 is the default)
%\MSC 0000 \sep 1111
\end{keyword}

\end{frontmatter}

%% \linenumbers

%% main text

\section{Introduction}
\label{sec:sample1}

Energy efficiency is essential for edge devices in resource-constrained smart environments, where reducing energy consumption is critical for extending device life and lowering maintenance costs \cite{tekin_review_2024}. Although techniques like network condition estimation and transmission power adjustment have been explored to improve energy efficiency at the network edge, optimizing deep neural network (DNN) models for on-device AI applications remains underexplored \cite{rahmani_improvement_2024}.
DNNs have proven to be highly effective for solving complex problems in smart environments, leveraging advancements from various fields, such as computer vision \cite{chai2021deep}. As Internet of Things (IoT), smart homes, and edge computing devices become more widespread, there is an increasing need to run DNN models directly on edge devices to enable real-time processing and minimize latency \cite{bimpas_leveraging_2024}. However, deploying large DNNs on these devices presents significant challenges due to their limitations, such as low-power processors and limited memory, making it difficult to manage intensive computations and large storage needs \cite{qi_small_2025}. Given that these devices typically operate on batteries, the high energy demands of DNNs can rapidly drain their power, reducing their operational longevity.

%\textcolor{blue}{Energy efficiency is crucial for edge devices operating in resource-constrained smart environments.  Minimizing energy consumption is key to extending device lifespan and reducing maintenance costs. While techniques such as network condition estimation and transmission power adjustment have been investigated to enhance energy efficiency at the edge of the network, the optimization of deep neural networks (DNNs) models for on-device AI applications remains underexplored \cite{rahmani_improvement_2024}.}
%DNNs have become highly effective tools for tackling complex challenges in smart environments, drawing on approaches from various fields, like computer vision \cite{chai2021deep}. The increasing deployment of IoT, smart homes and edge computing devices necessitates running DNN models directly on edge devices to enable real-time processing and reduce latency []. However, deploying large DNNs on resource-constrained devices is challenging due to their inherent limitations, such as low-power processors and limited memory, which make it difficult to handle intensive computations and large storage requirements. These devices rely on battery power, and the high energy consumption of DNNs can quickly deplete their batteries, reducing their operational lifespan.

To address these limitations, one strategy is to compress DNN models into smaller ones. Smaller models, with fewer parameters and layers, require less computational power and energy, making them more suitable for resource-constrained devices. However, their limited capacity can restrict their ability to capture complex patterns and nuances in data, potentially leading to lower accuracy on challenging tasks \cite{ma2019survey}. In contrast, large DNN models, with more parameters and layers, have a greater capacity to learn intricate features and relationships, often resulting in superior performance on complex problems. This increased complexity, however, demands substantial computational resources and memory \cite{yang2017method}, posing a significant challenge for deployment on edge devices.  

Efforts to mitigate the computational and memory burdens of DNNs have led to numerous works in the area of DNN compression \cite{neill2020overview}. These aim to reduce the model's computational and memory demands without compromising its performance. However, in practice, these efforts typically involve a trade-off between reducing model size and computational resources against the potential loss in model performance and predictive accuracy \cite{brownlee2021exploring}. 

A wide range of smart environment applications, from homes to cities, are driven by advancements in the field of computer vision \cite{gonzalez_garcia_midgar_2017}.
In computer vision, several methods have been developed to build efficient CNNs \cite{goel2020survey}. Quantization reduces the precision of weight values to conserve memory. Pruning involves the removal of redundant or insignificant neurons to decrease the computational load \cite{ECCLES202443}. Convolutional filter compression and matrix factorization reduce  model size by optimizing the structure of the network. Network Architecture Search (NAS) is another approach that finds DNNs optimized for individual devices, ensuring guaranteed performance. Additionally, knowledge distillation involves training a smaller \textit{student} neural network to mimic the behavior of a larger \textit{teacher} model, with the goal of maintaining performance while reducing model size \cite{tsanakas_light-weight_2024}.
All these approaches have several limitations, including a loss of accuracy in the compressed model, inefficient capturing of the teacher model's knowledge, and often considerable computational requirements, increased complexity and time needed for training, fine-tuning, and knowledge distillation \cite{violos_towards_2024}.

% What lead us to our methodology and hetergogeneity explanation
Our goal is to implement a CNN architecture that achieves the performance of a large deep CNN while maintaining the energy efficiency of smaller, more compact models. We explore deploying two small CNN models on resource-constrained devices in order to attain higher predictive accuracy  than each model could achieve individually. To this end, we investigate the concept of complementarity. Given a pair of small CNN models trained on the same dataset, complementarity is present if each model correctly predicts distinct subsets of the dataset. This suggests that each model has learned different aspects of the dataset. By leveraging the correct predictions from each model and discarding the incorrect ones, we aim to maximize the coverage of accurate dataset predictions.

To evaluate the validity of a prediction, we employed the confidence score based on the output logits of the model's prediction. This  provides a reasonably accurate estimate of prediction validity during inference, while having the least computational and energy requirements than other more advanced methods \cite{gawlikowski2022survey}. By utilizing this confidence score for both deployed models, we select the prediction that is deemed more accurate.

To further reduce energy consumption, we investigate the possibility of bypassing classification by invoking the CNNs only if the same input was not previously encountered. We focus on perceptual hashing \cite{du2020perceptual}, a technique that generates a hash value from an image's visual content, enabling similar images to have similar hash values. This method facilitates rapid identification and comparison of images. By storing the hash and classification for each previous prediction, we are able to avoid unnecessary CNN inferences.

%In this paper, we propose a DNN architecture aimed at reducing the energy requirements of inference by utilizing two Complementary CNNs augmented with a memory component. 
While our experiments are conducted on CNN architectures, we also refer to DNNs throughout the manuscript because CNNs are a specific subcategory of DNNs.
Our contributions include the following:

\begin{itemize}
  \item We propose using two complementary CNNs to improve accuracy by addressing each other's prediction weaknesses.
  \item We introduce a confidence-based mechanism that dynamically selects the more accurate CNN prediction.
  \item We present a memory system that stores previous classifications, allowing the system to bypass the more energy-intensive CNNs.
  \item We propose an architecture that integrates memory with inference, activating its components progressively to minimise energy consumption and response time.
\end{itemize}

%\item We introduce the concept of selecting a pair of complementary CNNs to maximize accuracy by complementing each other's weaknesses. \item We propose leveraging the prediction confidence of both CNNs to determine which prediction is more accurate.\item We present a memory component that records previous classifications for the same input, thereby allowing the system to bypass the more energy-intensive CNNs.

The structure of this paper is as follows: Section 2 reviews related work, Section 3 describes our proposed methodology, Section 4 reports our experimental results, and Section 5 presents our conclusions.

\section{Related Work}
The objective of our research is to reduce the energy consumption and response time of DNN inference, making DNN solutions suitable for resource-constrained edge devices without compromising prediction accuracy. As we will show in this section, the literature reveals a gap in solutions that can run on a single edge device, are not hardware-specific, and do not rely exclusively on compression techniques. 
Although compression techniques are the go-to approach, they involve a computationally intensive process like knowledge distillation or fine-tuning of pruned models, and they often result in significant performance degradation.

%To reduce the energy requirements, size, complexity, and inference time of DNNs, model compression has been proposed. 
Apart from pruning and knowledge distillation, the other two prominent compression approaches are quantization, and low-rank factorization \cite{neill2020overview}. These compression methods do not require a computational heavy process but they still result in a performance degradation. Recently, researchers have been investigating approaches such as automating the deactivation of DNN layers, offloading computational workloads to other computing nodes, and co-designing hardware and software. We will briefly present these approaches along with their limitations to better contextualize our work within the relevant literature.

In addition to pruning and knowledge distillation, two other widely used compression techniques are quantization and low-rank factorization \cite{neill2020overview}. While these methods are computationally efficient, they still lead to performance degradation. Recently, researchers have been investigating approaches such as automating the deactivation of DNN layers, offloading computational workloads to other computing nodes, and co-designing hardware and software. We will briefly present these approaches along with their limitations to better contextualize our work within the relevant literature.

%More recently, researchers have explored alternative strategies such as automating the deactivation of DNN layers, offloading computational tasks to other nodes, and jointly designing hardware and software for greater efficiency. In this section, we will provide an overview of these approaches and their limitations to better situate our work within the current research landscape.

%The four prominent compression approaches are pruning, knowledge distillation, quantization, and low-rank factorization \cite{neill2020overview}. While these methods significantly decrease the size and energy consumption of DNNs, they also often result in reduced model performance. Recently, researchers have been investigating approaches such as automating the deactivation of DNN layers, offloading computational workloads to other computing nodes, and co-designing hardware and software. We will briefly present these approaches along with their limitations to better contextualize our work within the relevant literature.

%\begin{figure*}[ht!]
%\centering
%\includegraphics[width=\textwidth]{Figures/theirs.png}
%\caption {Baseline Model With Big and Little DNNs}
%\label{fig:expe1}
%\end{figure*}

\cite{hu2019dynamic} present a method aimed at improving the energy efficiency of DNN inference on edge devices by dynamically adapting the network structure in real time. This technique, known as \textit{adaptive DNN surgery}, selectively activates or deactivates certain layers and neurons based on the current computational load and resource availability. 

By doing so, the network can optimize its inference speed and reduce energy consumption, making it well-suited for energy-constrained edge environments. The dynamic adjustments ensure that only the necessary parts of the network are active at any given time, thus minimizing unnecessary computations and associated energy costs.
The limitation of this method is that it is trained for on a specific prediction model and for every new use case the process should be retrained to activate and deactivate the layers and neurons.

\cite{zeng2020coedge} present a method to enhance energy efficiency in DNN inference by offloading tasks across multiple complementary edge devices. CoEdge dynamically distributes the inference workload based on each device's computational capabilities and current energy levels, ensuring optimal resource utilization and minimizing overall energy consumption. 
Real-time adaptability and dynamic load balancing enable efficient resource allocation. In this context, Adaptive Stochastic Learning Automata can dynamically distribute resources to achieve optimal performance \cite{wang_load-aware_2024}. Additionally, Multi-Agent Deep Learning has been proposed to adjust resource allocation and enhances the performance of processing nodes \cite{kuang_reliable_2024}.
These "cooperative" approaches prevent any single device from becoming overburdened, extending the battery life of all devices involved.

\cite{zhang2024coarse} introduce the \textit{coarse-to-fine} method to enhance the efficiency of DNN inference on resource-constrained edge devices. This operates by initially deploying a lightweight, coarse-grained model to make fast, preliminary predictions. When the confidence score, also known as the confidence level, of these initial predictions is sufficiently high, the results are accepted. If the confidence score is low, indicating uncertainty, the system escalates the task to a more complex, fine-grained model for further analysis and refined predictions. This tiered approach ensures that only a fraction of the tasks incur the heavy computational load of the large model, thus optimizing the use of limited edge resources. By performing initial coarse analysis on edge devices and offloading expensive analysis to more powerful servers if needed, coarse-to-fine optimizes resource utilization across the edge network. 

While the two above task offloading methods provide improvements in energy requirements they rely on a group of devices on the edge network or to the cloud for the more demanding tasks. Our work surpasses this limitation because it enhances the performance on a single device for all possible inference tasks without burdening other computing devices.

Other studies address the DNN optimization problem through hardware/software co-design. This is an integrated approach that involves jointly designing both the hardware and software components of a system to optimize performance and energy efficiency.
\cite{hao2019fpga} introduce a method to improve energy efficiency and performance of DNNs on edge devices by co-designing both hardware (FPGA) and software DNN. The approach leverages the reconfigurable nature of FPGAs to create customized, energy-efficient hardware accelerators tailored specifically for DNN tasks. It also employs algorithmic techniques such as model compression and quantization to optimize the DNNs for reduced computational load. This co-design ensures that both the hardware and software are highly tuned to work together, significantly lowering energy consumption and enhancing processing speed.

\begin{figure*}[ht!]
\centering
\includegraphics[width=\textwidth]{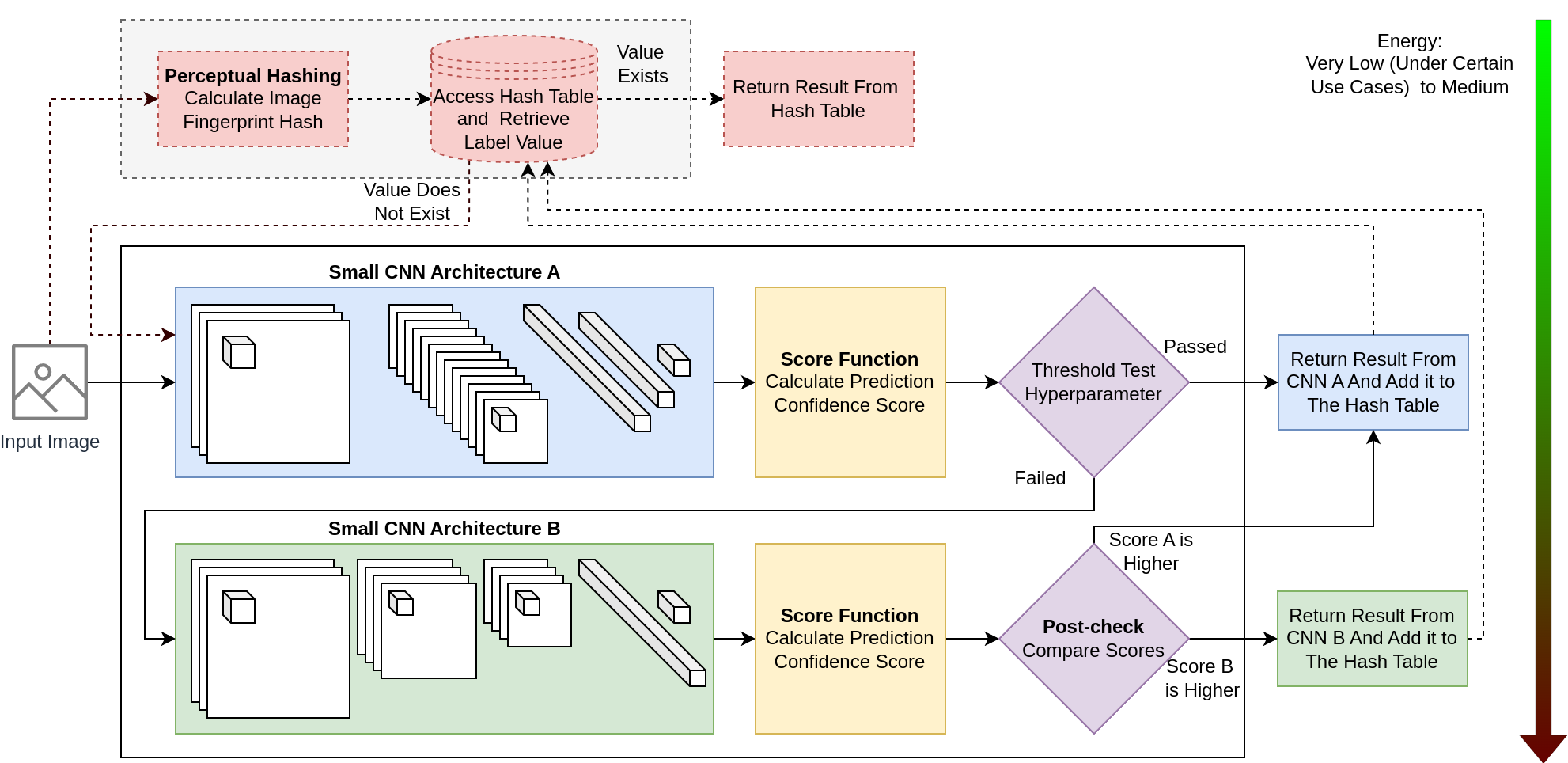}
\caption {Overview of our proposed methodology, which consists of two complementary small CNN's with a memory component.}
\label{fig:expe2}
\end{figure*}

\cite{lee2020hardware} outline a strategy to enhance the energy efficiency of DNN processors on mobile devices through the integrated optimization of hardware and algorithms. By designing specialized, low-power processors and employing algorithmic techniques like model compression, quantization, and pruning, this approach reduces computational demands while maintaining high accuracy. The co-design ensures mutual optimization, significantly cutting down energy consumption and making sophisticated DNN models feasible for real-time, on-device artificial intelligence applications.

The research limitation we address, compared to the two approaches mentioned above, is that we aim to provide a general method for improving the inference energy requirements of DNNs in a hardware-agnostic manner, with minimal model optimizations or modifications. Our goal is to enable rapid deployment on a single device using already trained DNN models.

The closest research to our work was presented by \cite{park_biglittle_2015}. This employs a dual network architecture that dynamically switches between a "small" CNN for simpler tasks  and a "big" CNN for complex tasks based on whether the confidence score of the small CNN falls below a certain threshold. %, as shown in Figure \ref{fig:expe1}. 
This method significantly reduces energy consumption during inference without sacrificing accuracy. The idea of using two CNN models and selecting the most suitable prediction for a specific input inspired us to explore how two network architectures can collaborate effectively. Our paper distinguishes itself from this work by making further research contributions through complementarity, post-check mechanism, and perceptual hashing. Additionally, a limitation of the work by \cite{park_biglittle_2015} is that they used a simulator for their experimental evaluation - in particular, they rely on a Verilog-emulated hardware accelerator - whereas we conducted our experiments on an actual edge device and a power meter.

A core concept we use to enable efficient collaboration between the two models is based on the confidence score. \cite{Jayakodi_2018}  introduced a method for calculating a confidence estimate of a model's prediction. Specifically they present a method to balance the accuracy and energy consumption of DNN inference on embedded systems. %Their proposed co-design approach involves jointly optimizing both the DNN model and the hardware to achieve an efficient trade-off. 
We applied this method to decide which DNN will provide the prediction outcome.

% A core concept we use to enable efficient collaboration between the two models is based on the confidence score and the quantification of prediction uncertainty. \cite{gawlikowski2022survey} provide a comprehensive survey of various methods addressing prediction uncertainty during both training and inference. One of the methods discussed in their paper involves calculating prediction uncertainty estimates based on the output of a DNN during a single inference. We favored this method due to its minimal computational and energy requirements. Building on this foundation, \cite{Jayakodi_2018} introduced a technique for estimating the confidence of a model's prediction. Specifically, they present a method to balance the accuracy and energy consumption of DNN inference on embedded systems. Their proposed co-design approach involves jointly optimizing both the DNN model and the hardware to achieve an efficient trade-off. We applied this method to determine which DNN would provide the prediction outcome, further enhancing our system's efficiency and performance.

For the function of our memory component we explored the use of perceptual hashing. \cite{tang2012perceptual} presents a method for generating perceptual hashes for color images based on invariant moments. %Invariant moments are mathematical features that remain unchanged under image transformations such as scaling, rotation, and translation. 
The proposed technique computes these moments from the color image to create a unique hash value that reflects the image's visual content. %As we will present in our experimental analysis, 
We empirically found that these moments are not fully invariant, i.e. they slightly vary under various transformations. A further improvement of the invariant moments is presented in the work of \cite{flusser2009moments}, where they implement invariants from complex moments. These complex invariants are fully invariant to transformations, and as such we adopted them in our methodology.

While model compression techniques such as pruning, knowledge distillation, and quantization aim to create a smaller model that replaces a larger one, our approach is fundamentally different. Instead of focusing on compressing a single model, our work emphasizes the effective collaboration between two complementary small models, each with strong accuracy and confidence in distinct parts of the data samples. This approach is more effective than dynamic switching between a small and large model, as the latter often relies on a large model that may not be suitable for edge devices. Additionally, our method incorporates a memory component, which is absent from other techniques in the literature. This memory component works in synergy with the inference process. Our proposed method selectively activates the memory and inference components as needed, optimizing energy consumption and response time. This approach further distinguishes our work from traditional compression methods.

\section{Proposed Methodology}
\label{Sec:ProposedMethodology}
Our proposed methodology leverages two complementary CNNs to optimize prediction performance while minimizing computational and energy demands. These CNNs operate sequentially: for a given input, the first CNN is initially invoked. The output of the first CNN is evaluated by a prediction confidence score based on the CNN's logits vector. This score is then compared against a threshold, and if it falls short, the second CNN is invoked. The threshold is a predefined value that determines the extent of usage of the second CNN as described in subsection \ref{SubSec:MethodologyThresh}.
For the second CNN, a prediction confidence score is similarly calculated using the score function. If both CNNs are utilized for a prediction, their respective confidence scores are compared, and the final decision is based on the more confident prediction. % with the higher confidence score. 

We also employ a memory component designed to store prior classifications for identical or highly similar inputs. When an input first passes through the memory component, a unique image fingerprint is generated. This fingerprint is used as a key to access a hash table. If the hash table contains a value for the key, %corresponding to the classification of the input, 
this is returned without invoking the CNNs. If no value exists, indicating a new input, the prediction process as described above is initiated. After classification using the CNNs, the classification label is saved to the hash table with the image fingerprint as the key. This component is illustrated in the upper part of Figure \ref{fig:expe2}, enclosed in a dashed box. Note that the proposed dual-CNN architecture can also operate without this component if it is considered preferable in specific application contexts.

Figure \ref{fig:expe2} illustrates an overview of our solution’s architecture, which comprises three main components: the small CNN architecture A, the small CNN architecture B, and the memory component. These components are activated based on score comparisons.
When an input image arrives, it is first processed by the memory component, specifically the perceptual hashing module, to generate a fingerprint hash of the image. If this fingerprint hash already exists in the access hash table, the corresponding label is retrieved, and the result is returned from the hash table. If the hash is not found, the image is passed to the small CNN architecture A, which makes a prediction and computes a confidence score. If this score exceeds a predefined threshold, the prediction is returned and the hash table is updated. Otherwise, the image is forwarded to the small CNN architecture B, which also generates a prediction and a confidence score. The confidence scores are compared in a post-check, and the prediction with the highest score is returned, while the hash table is updated accordingly.

An important aspect, shown by the energy arrow on the right of Figure \ref{fig:expe2}, is the varying energy consumption. If the image already exists in the hash table, the energy usage is minimal. When the image is processed by the small CNN architecture A, the energy increases, and it rises further if the small CNN architecture B is also engaged. The following subsections describe each component and the checks of the proposed solution in more detail.

\subsection{Two Complementary CNNs}
\label{SubSec:ProposedMethodology_2Het}
Our methodology incorporates two distinct and compact CNN architectures. This strategy capitalizes on the diverse design principles of each model, allowing them to complement each other's deficiencies. In instances of classification ambiguity, the utilization of an alternative CNN mitigates such limitations, given that each neural network captures unique facets of the dataset's information. Opting for two small models, as opposed to a mixture of large and small ones, further minimizes power consumption as we will see in the experimental evaluation.

% The concept of complementarity is illustrated in Figure \ref{fig:het-demo}. The labels of a given dataset represented by gray boxes, and the correct predictions made by a pair of CNNs represented by blue and green circles as subsets of the dataset. The complementarity is function of the symmetric difference between the correct predictions of the two CNNs. The four parts of the Figure \ref{fig:het-demo} shows four examples that help us to understand intuitively the idea of complementarity.

\begin{figure}[ht!]
\centering
\includegraphics[width=0.5\textwidth]{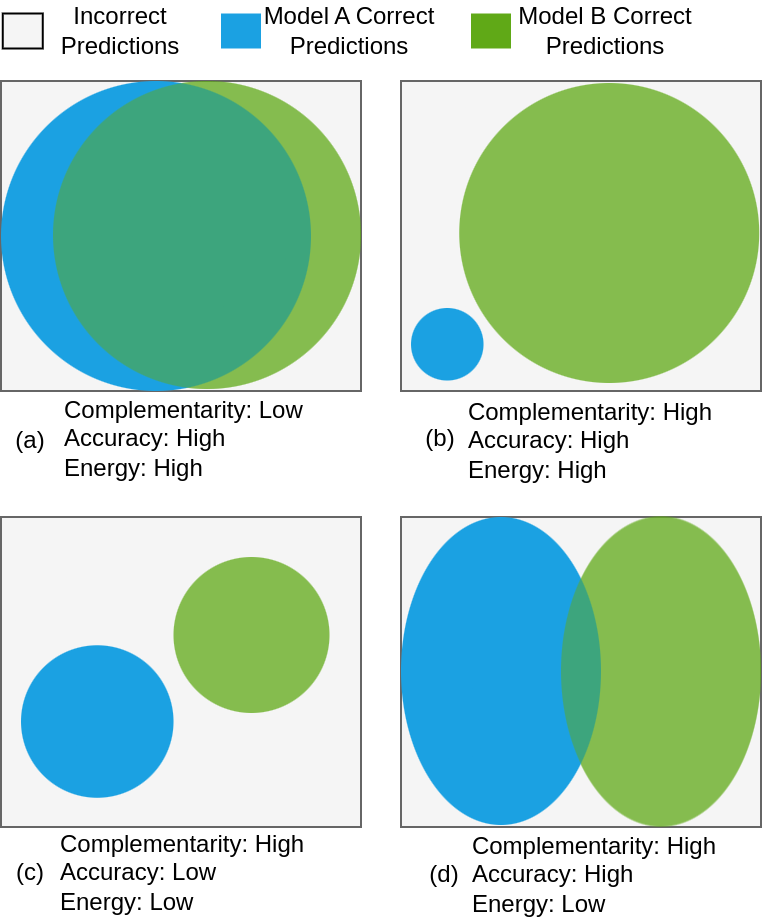}
\caption {Complementarity based on the predictions of two CNN models.}
\label{fig:het-demo}
\end{figure}

The concept of complementarity is illustrated in Figure \ref{fig:het-demo}. The gray box represents the whole dataset, of which we consider that the volume outside the circles represent the percentage of labels that are incorrectly predicted by both models and the correct predictions made by a pair of CNNs represented by blue and green circles as subsets of the dataset. The complementarity is function of the symmetric difference between the correct predictions of the two CNNs. The four parts of the Figure \ref{fig:het-demo} shows four examples that help us to understand intuitively the idea of complementarity.

In the case of Figure \ref{fig:het-demo}a, both models correctly predict a large portion of the dataset, but their predictions overlap significantly. This substantial intersection indicates low complementarity, as the models frequently predict the same samples. Consequently, this results in redundant resource utilization and increased energy consumption. 

Figure \ref{fig:het-demo}b depicts a scenario with high complementarity and extensive correct prediction coverage. However, the size disparity between the models is not optimal, as the larger green model requires higher energy consumption. 

Figure \ref{fig:het-demo}c illustrates a fully complementary pair with no overlap in their correct predictions. However, their overall prediction coverage is low, resulting in limited accuracy. Despite their perfect complementarity and equal size, the limited complexity and small size of these models lead to suboptimal accuracy.

Finally, Figure \ref{fig:het-demo}d presents the ideal scenario, where both models have similar size, and together they cover a large portion of the dataset with minimal overlap, thus achieving high complementarity and optimal resource utilization.

%In order to quantify the complementarity of a selected pair of DNN models, we present the following formula:

We define the complementarity of a pair of CNN models with the formula given in Eq. \ref{complementarity}

% before
% \begin{equation} \label{complementarity}
% complementarity(A, B) = \frac {A \cup B - 2 \cdot A \cap B - \left| |A| - |B| \right| } {N}
% \end{equation}
% where $A$ and $B$ are the number of true predictions of models A and B and $N$ the number of samples in a given dataset. 

%after

\begin{equation} \label{complementarity}
complementarity(a, b) = \frac {n(a \cup b) - n(a \cap b) - \left| n(a) - n(b) \right| } {N}
\end{equation}

\medskip
Given all the correct predictions from a dataset as represented by the set $N$, then the subsets $a$ and $b$ represent the correct predictions of models a and b respectively.

\medskip
The first term is defined as:

\begin{equation} \label{eq:term_1_definition}
n(a \cup_{}^{} b) = \sum_{i=1}^{N} \mathbb{I} ^ 1(y_i^a, y_i^b, y_i^{true})
\end{equation}
where the indicator function $\mathbb{I}^1$ is defined as:

\begin{equation} \label{eq:term_1_indicator}
\mathbb{I}^1 (y^a,  y^b, y^{true}) = \begin{cases} 1, & \text{if } y^a = y^{true} 
\\ 1, & \text{if }  y^b = y^{true} \\0, & \text{otherwise} \end{cases}
\end{equation}

The second term is defined as:

\begin{equation} \label{eq:term_2_definition}
n(a \cap_{}^{} b) = \sum_{i=1}^{N} \mathbb{I} ^ 2(y_i^a, y_i^b, y_i^{true})
\end{equation}
where the indicator function $\mathbb{I}^2$ is defined as:

\begin{equation} \label{eq:term_2_indicator}
\mathbb{I}^2(y^a, y^b, y^{true}) = \begin{cases} 1, & \text{if } y^a = y^{true} \text{ and } y^b = y^{true} \\ 0, & \text{otherwise} \end{cases}
\end{equation}

The third term components are defined as:

\begin{equation} \label{eq:term_3a_definition}
n(a) = \sum_{i=1}^{N} \mathbb{I}^3(y_i^a, y_i^{true})
\end{equation}

\begin{equation} \label{eq:term_3b_definition}
n(b) = \sum_{i=1}^{N} \mathbb{I}^4(y_i^b, y_i^{true})
\end{equation}

where the indicator functions $\mathbb{I}^3$ and $\mathbb{I}^4$ are defined as:
\begin{equation} \label{eq:term_3a_indicator}
\mathbb{I}^3(y^a, y^{true} ) = \begin{cases} 1, & \text{if } y^a = y^{true}  \\ 0, & \text{otherwise} \end{cases}
\end{equation}

and

\begin{equation} \label{eq:term_3b_indicator}
\mathbb{I}^4(y^b, y^{true} ) = \begin{cases} 1, & \text{if } y^b = y^{true}  \\ 0, & \text{otherwise} \end{cases}
\end{equation}

respectively.

\medskip
For the above equations \ref{eq:term_1_definition}, \ref{eq:term_2_definition}, \ref{eq:term_3a_definition} and \ref{eq:term_3b_definition}, $y_i^a$ is the prediction from model a, $y_i^b$ is the prediction from model b and $y_i^{true}$ is the true label for a given input $i$.
\color{black}
\medskip
%Before
% The formula consists of three terms in the numerator, each representing different aspects of model performance.
% The first term, \( A \cup B \), represents the total number of correct predictions covered by either of the two models in the given dataset. The objective is to maximize this term to ensure that the combined predictive capability of the models is as comprehensive as possible.

% The second term, \(A \cap B\) , denotes the intersection of correct predictions made by both models. The goal is to minimize this term. Minimizing the overlap of correct predictions between the models reduces redundancy, thereby conserving computational resources. Ideally, we aim to utilize smaller models with minimal overlap that collectively cover the same prediction space as larger models.

% The third term, \( \left| n(A) - n(B) \right| \), measures the disparity in the sizes of the two models. This term should also be minimized. A case where a significant size disparity exists, where one large model may cover the majority of predictions (e.g., 95\%), while a much smaller model covers the remainder (e.g., 5\%), could also be considered a complementary pair. However, this would lead to inefficient resource usage, as it relies heavily on a large, resource-intensive model. Ideally, the models should be of comparable size, each covering distinct portions of the dataset’s predictions to promote balanced and efficient use of computational resources.

% After
The formula (\ref{complementarity}) consists of three terms in the numerator.
The first term, $ n(a \cup b) $, represents the total number of correct predictions covered by either of the two models in the given dataset. The objective is to maximize this term to ensure that the combined predictive capability of the models is as comprehensive as possible.

%paraphrased)
The second term, $n(a \cap b)$, represents the overlap of correct predictions between the two models, meaning, the instances that both models predict correctly. The objective is to minimize this overlap since it is subtracted from the complementarity score. The rationale is that by reducing the intersection of correct predictions helps eliminate redundancy, optimizing the use of computational resources. Ideally, the goal is to use smaller models with minimal overlap, ensuring they together cover the same prediction space as larger models.
The two models become accurate on different data subsets, covering each other’s weaknesses without leading to poor generalization. The models' capacity to generalize is preserved because they do not overfit, neither on samples where they exhibit high confidence nor on samples where they exhibit low confidence.

The third term, $| n(a) - n(b) |$, measures the disparity in the sizes of the two models in terms of number of correct predictions. This term should also be minimized. A case where a significant size disparity exists, where one large model may cover the majority of predictions (e.g., 95\%), while a much smaller model covers the remainder (e.g., 5\%), could also be considered a complementary pair. However, this would lead to inefficient resource usage, as it relies heavily on a large, resource-intensive model. Ideally, the models should be of comparable size, each covering distinct portions of the dataset’s predictions to promote balanced and efficient use of computational resources.

%Figure \ref{fig:complementarity} demonstrates the idea.
% not sure about this
% \begin{figure}[ht!]
% \centering
% \includegraphics[width=0.5\textwidth]{Figures/complementarity demonstation.drawio.png}
% \caption {A demonstration of the complementarity idea}
% \label{fig:complementarity}
% \end{figure}

\subsection{Confidence Score Functions}
\label{SubSec:confidence_score_functions}
CNNs in a classification problem produce a logits vector $\vec{z}$, which, upon passing  through a softmax function (\ref{softmax}), is converted into a probability distribution vector $\vec{p}$. 

\begin{equation} \label{softmax}
\sigma (\vec{z})_i = \frac{e^{z_i}} {\sum_{j=1}^{N}e^{z_j}}
\end{equation}
where $i$ and $j$ are the $i$-th and $j$-th element of the logits vector.

\medskip
Each dimension of this vector corresponds to a class, with the value $\sigma (\vec{z})_i$ indicating the probability that the input belongs to that class. We then operate on this probability vector applying one of three score functions, as presented in \cite{Jayakodi_2018}:

\medskip
a) The Max Probability score function (\ref{max_P}) simply selects the highest value from the probability distribution vector. When the CNN is confident in its prediction, this top probability will approach 1.
\begin{equation} \label{max_P}
score(\vec{p}) = max(\{p_1,...p_n\})
\end{equation}

\medskip
b) The Difference score function (\ref{difference}) calculates the disparity between the first and second highest values in the output probability vector. 

\begin{equation} \label{difference}
score(\vec{p}) = p_i - p_j
\end{equation}
where $p_i$ is the largest value in the probability vector $\vec{p}$ and $p_j$ is the second largest value. A larger difference signifies greater confidence in the prediction, as the highest value approaches 1 while the second highest approaches 0. 
The Difference score function sometimes outperforms the Max Probability in confident predictions, as the top probability may be high, but the second may also hold a small yet significant value.

\medskip
c) The Entropy score function (\ref{entropy-norm}) computes the entropy of the output probability vector. A lower entropy value corresponds to higher prediction confidence. 
In theory, lower entropy suggests higher confidence in prediction, indicating that the CNN has confidently assigned a single class. Conversely, higher entropy suggests a more evenly spread distribution, indicating lower confidence.

\begin{equation} \label{entropy}
\text{entropy} = -\sum_{i=1}^{N} p_{i} \ln (p_{i})
\end{equation}

We employed a normalized version of the standard entropy formula to convert the score values into the range of $[0, 1]$, making them more comparable to the other two score functions:
\begin{equation} \label{entropy-norm}
score(\vec{p}) = \frac {-\sum_{i=1}^{N} p_{i} \ln (p_{i})}{-\sum_{i=1}^{K} \frac{i}{K} \ln \left (\frac{i}{K}  \right )}
\end{equation}
where $K$ is the total number of classes of the dataset.

\medskip
The selection of a score function among the three available options depends on the specific CNN pair under consideration, as the performance of each function can vary across different pairs. Empirically, we have determined that the Difference score function (\ref{difference}) generally yields the best performance in the majority of cases. This observation aligns with the findings reported by \cite{park_biglittle_2015}.

\medskip
\subsection{Score Comparison \& Post-check}
\label{SubSec:ProposedMethodology_PostCheck}
%The calculated score value of the selected scoring function is utilized in two distinct manners. 
The confidence score is utilized in two steps.
First, it is employed to compare the score value of the initial CNN against a predetermined threshold. This comparison determines whether to trigger the subsequent CNN. Second, after the invocation of the second CNN, a second confidence score is computed. The two confidence scores are compared, and the prediction of the CNN with the highest (or lowest, if the entropy score function is applied) score is selected. We name this comparison ``post-check'', indicating an assessment that follows the inference and confidence score of the second CNN.
This approach is advantageous as we will see in the experimental evaluation because in some cases the initial CNN may show greater confidence and potentially higher accuracy in its predictions than the second CNN, even if it fails to surpass the threshold test.

\subsection{Threshold Hyper-parameter}
\label{SubSec:MethodologyThresh}
The threshold hyperparameter is a fixed value that determines the extent of usage of the second CNN. By employing lower values (or higher in the case of the entropy score function), the utilization of the second CNN is reduced, thereby decreasing energy consumption but also affecting accuracy. However, by employing a higher threshold value, although the invocation of the second CNN becomes more frequent, this does not necessarily result in higher accuracy, as evidenced by \cite{park_biglittle_2015}. 

Our selection of CNN architectures is primarily influenced by the complementary nature of the two CNNs. Since both are nearly equal in size, prediction accuracy, and capabilities, there is no inherent advantage of the second CNN over the first; however, since the first CNN is used 100\% of the time, it is preferable to refrain from using the second as much as possible. Nonetheless, there exists an optimal trade-off that could maximize accuracy. In our methodology, we compute the value that leads to maximum accuracy for the selected CNNs architectures by using the following method in the training dataset.

If we define the achieved accuracy result of our setup, given $N$ samples and a $\lambda$ value as:
% \begin{equation}\label{eq:acc}
%    acc(\lambda) = \frac{1}{N} \sum_{i=1}^{N} \begin{cases} 1, & \text{ if }\ {H(a,b,\lambda,s) }= y^{true}\\ 0, & \text{ otherwise }
% \end{cases}
% \end{equation}
\begin{equation}\label{eq:acc}
   acc(\lambda) = \frac{1}{N} \sum_{i=1}^{N} \begin{cases} 1, & \text{ if }\ {H(a(x_i), b(x_i), s, \lambda) }= y_i^{true}\\ 0, & \text{ otherwise }
\end{cases}
\end{equation}
where the function $H$ represents our methodology setup that returns a prediction $\hat{y}$ in a single inference and $a(x_i)$ and $b(x_i)$ are the predictions of model $a$ and model $b$ respectively of the selected CNN pair given an input $x_i$, $s=score(\vec{p})$ the selected of the three scoring functions and $\lambda$ the provided threshold hyper-parameter. The optimal value of $\lambda$, denoted as $\lambda^*$, is determined as the one that maximizes the accuracy of the selected setup and can be calculated:

\begin{equation} \label{find-best-lambda}
\lambda^* = \argmax_{\lambda} (acc(\lambda)),\ \text{ for }\ 0 < \lambda < 1
\end{equation}
where $acc(\lambda)$ is the equation (\ref{eq:acc}).

\medskip
%Before
% By running Equation (\ref{find-best-lambda}) for each of the three score functions, Max probability (\ref{max_P}), Difference (\ref{difference}) and Entropy (\ref{entropy-norm}), we can select the combination of score function and $\lambda^*$ parameter, that achieves maximum accuracy for a selected \textcolor{blue}{ANN pair}.
%After

To determine the optimal threshold hyperparameter, we first select a pair of CNNs exhibiting a high complementarity score based on equation (\ref{complementarity}) and one of the confidence score functions (subsection \ref{SubSec:confidence_score_functions}). We then apply our methodology iteratively for various $\lambda$ values within the range $[0, 1]$. This procedure is repeated for each score function, allowing us to identify the optimal combination of score function and $\lambda^*$ that maximizes the accuracy for the chosen CNN pair.

\begin{figure}[ht!]
\centering
\includegraphics[width=0.4\textwidth]{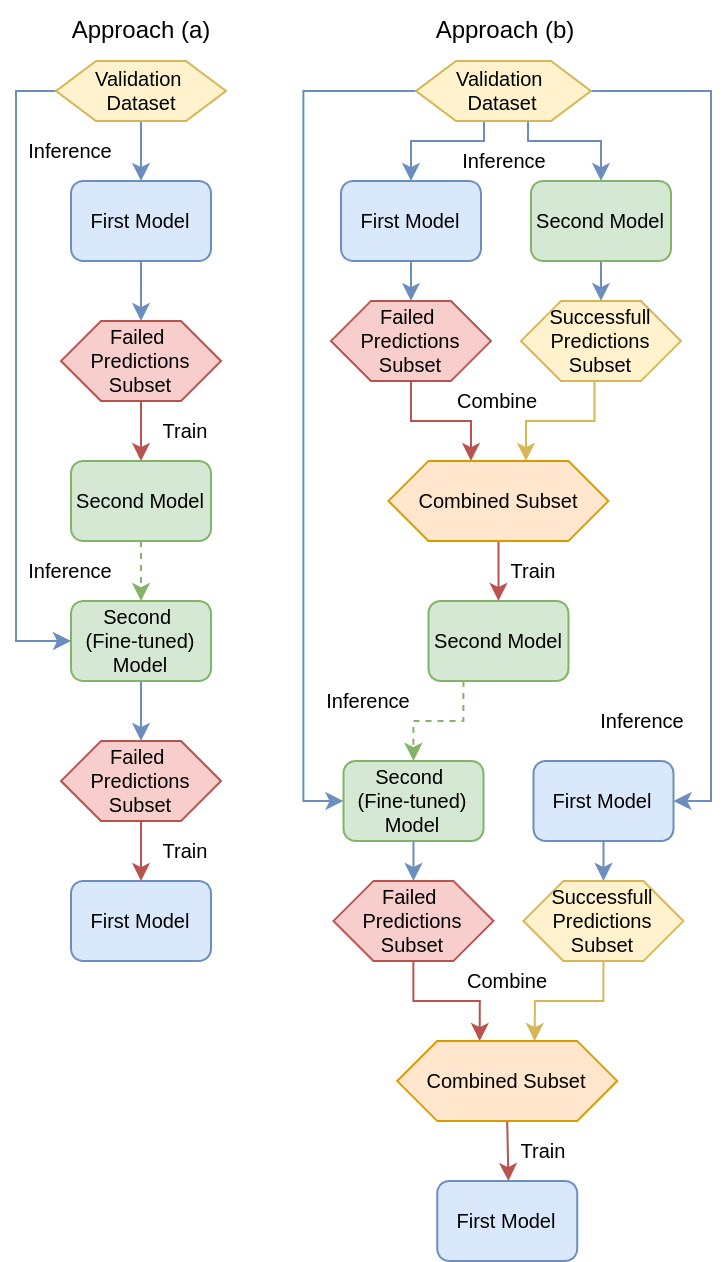}
\caption {Fine-tuning to increase complementarity.}
\label{fig:fine-tuning}
\end{figure}

\subsection{Memory Component}
\label{SubSec:ProposedMethodology_Mem}
% Original
%As an enhancement to our CNN complementary methodology, we incorporate a memory component designed to reduce energy consumption during predictions under specific conditions. This component aims to recall whether a previous classification has been made for a given input, thereby bypassing the need to invoke the CNN when possible. To implement this, we utilize Perceptual Hashing \cite{tang2012perceptual}, a technique for generating a unique fingerprint based on the contents of an image.

% Modified
As an enhancement to our CNN complementary methodology, we incorporate a memory component designed to reduce energy consumption during predictions under specific conditions. This component aims to recall whether a previous classification has been made for a given input, thereby bypassing the need to invoke the CNN when possible. As noted by \cite{yang2017method}, data movement through memory is a highly energy-intensive process. To address this, we utilized a combination of perceptual hashing \cite{tang2012perceptual}, which generates a unique fingerprint based on the contents of an image, and a hash table data structure to store and retrieve classification labels for each input. For each input, the required memory access operations are either one read, if the image classification label exists in the hash table, or one read and one write operation, if the input image is new. This approach ensures that energy consumption is kept to a minimum.
% End modification

% Maybe this paragraph needs to be rewritten aswell as a continuation of the above.
According to Figure \ref{fig:expe2}, every new input image first passes through the memory component, where a hash value is calculated for that image. This is then used as a key to either read or add it, if it does not exist, to a key-value store. The value corresponds to the classification label of the image. If the same image has been classified before, then the read operation using the calculated hash value from the hash table will return the class label, thus skipping the invocation of the CNNs. If no such key exists in the store, then the image is a new input and is thus provided to the CNNs for classification. After the classification, the label, along with the calculated image hash, is added to the key-value store.

To calculate the image hash code, we explored two Perceptual Hashing methods \cite{flusser2009moments}: 

\begin{enumerate}
    \item Difference Hash 
    \item Invariants from Complex Moments.
\end{enumerate}

The Difference Hash method is the simplest and least computationally intensive approach. It functions by detecting gradients within the image. A bit sequence is computed wherein each bit and denotes a change in brightness between adjacent pixels. The setting of each bit is determined by whether the brightness of the left pixel surpasses that of the right one. This sequence is subsequently serving as the image fingerprint and key for the hash map. Although the Difference Hash method offers rapid and energy-efficient hashing, it is susceptible to alterations in the image, including significant fluctuations in brightness and contrast, as well as rotations and mirroring, resulting in distinct calculated fingerprints.

The second method utilizes complex moments of an image to calculate fully rotation-invariant features that remain unchanged despite spatial transformations, as described in \cite{flusser2009moments}. We obtain six invariant features represented as a six-dimensional vector of floating points. %Of the six floating-point values, 
All of these values are fully invariant to rotations, and four of them are both fully invariant to rotations and additionally to mirroring in any axis, meaning that even if the image is rotated or mirrored these four values remain the same. We sum these four fully invariant values to receive a single floating-point number that represents the image and is considered the fingerprint, subsequently used as the key in the key-value store, as described above. %is more computationally heavy and 
% End of modification

% Perceptual hashing algorithms are designed to produce similar hashes for images with small changes, facilitating comparison with a database of saved hashes to find the most similar match. This method requires frequent memory access to address a significant search space. One of the most energy required operations is data movement through memory \cite{yang2017method}. However, due to our goal of reducing energy consumption, extensive memory access is not feasible for us, thus the memory component is applicable to identical images using the hash map method. While the whole process adds a slight computational overhead, our experimental results show a significant decrease in energy requirements when used under the conditions of multiple identical image inferences.

%NEW Fine-tuning
\subsection{Enhancing Complementarity}
\label{SubSec:IncreasingComplementarity}

To enhance the complementarity of a chosen pair, we fine-tune the CNN models.
%We experimented with fine-tuning the CNN models in a selected pair to enhance their complementarity. 
Our objective was to "push" each model toward different areas of the dataset, thereby improving accuracy gains and overall performance. The core idea was to use the failed prediction instances from one model to train the other model for a few epochs, and vice versa. 
We explored two primary approaches to fine-tuning: (a) using only the instances where predictions failed and (b) using both the failed and successful prediction instances.

In the first approach, shown on the left side of Figure \ref{fig:fine-tuning}, we select a complementary pair of models and pass the validation dataset through the first model. We then collect all instances where the model makes incorrect predictions into a separate subset. This subset is used to train the second model for a few epochs. After fine-tuning the second model, we reverse the process. We pass the validation dataset through the second, now fine-tuned model, collect the failed predictions into a separate subset, and use this subset to fine-tune the first model.

\begin{table*}[]
\centering
\begin{tabular}{@{}llccccccc@{}}
\toprule
\multicolumn{1}{c}{\multirow{2}{*}{\textbf{Dataset}}} &
  \multicolumn{1}{c}{\multirow{2}{*}{\textbf{Model}}} &
  \multirow{2}{*}{\textbf{\begin{tabular}[c]{@{}c@{}}N. Parameters \\ (M.)\end{tabular}}} &
  \multicolumn{3}{c}{\textbf{Response Times (ms)}} &
  \multirow{2}{*}{\textbf{\begin{tabular}[c]{@{}c@{}}Current\\ (mAh)\end{tabular}}} &
  \multirow{2}{*}{\textbf{\begin{tabular}[c]{@{}c@{}}Power\\ (Wh)\end{tabular}}} &
  \multirow{2}{*}{\textbf{\begin{tabular}[c]{@{}c@{}}Accuracy\\  (\%)\end{tabular}}} \\
\multicolumn{1}{c}{} & \multicolumn{1}{c}{} &       & \textbf{mean} & \textbf{95th} & \textbf{99th} &     &      &       \\ \midrule
CIFAR-10             & Resnet 20            & 0.27  & 22.3          & 23.3          & 23.8          & 48  & 0.25 & 92.60 \\
CIFAR-10             & MobileNetV2-0.5      & 0.70  & 42.4          & 43.8          & 44.3          & 81  & 0.41 & 93.12 \\
CIFAR-10             & RepVGG-A0            & 7.84  & 51.9          & 53.8          & 54.4          & 125 & 0.64 & 94.46 \\
CIFAR-10             & RepVGG-A2            & 26.82 & 64.8          & 65.3          & 67.0          & 225 & 1.20 & 95.27 \\
CIFAR-10             & ShuffleNet V2 X-1.0 (distilled)  & 1.26  & 53.0         &  54.4       &   56.1       & 40  & 0.75  & 77.36 \\
ImageNet             & MnasNet-1.3          & 6.3   & 46.6          & 47.2          & 85.1          & 165 & 0.88 & 69.17 \\
ImageNet             & DenseNet-121         & 7.9   & 131.0         & 141.1         & 149.0         & 398 & 2.12 & 69.65 \\
ImageNet             & RegNet-X-800MF       & 7.3   & 118.4         & 127.7         & 133.1         & 283 & 1.50 & 67.89 \\
ImageNet             & RegNet-X-8GF         & 39.6  & 235.1         & 237.8         & 238.7         & 845 & 4.50 & 73.11 \\

Intel           & MobileNetV2-X1.0  & 2.24  & 43.6  & 45.3         & 52.8       & 0.13         & 24 & 86.70 \\ 
Intel           & Resnet 44  & 0.66   & 44.6  & 47.2 & 48.2       & 0.11         & 20 & 85.60\\
Intel           & RepVGG-A0  & 7.84  & 54.4  & 55.2         & 56.3      & 0.16         & 30 & 81.83 \\ 
Intel           & RepVGG-A2  & 26.82   & 65.0  & 65.5 & 87.6      & 0.28        & 54 & 88.03\\ 
Intel           & ShufflenetV2-X1.5 (distilled)  & 2.48   & 49.3  & 54.0 & 55.6      & 0.12       & 23 & 82.20\\

FashionMNIST           & MobileNetV2-X1.0  & 2.24  & 44.8  & 46.8         & 47.1      & 0.34         & 65 & 86.86 \\ 
FashionMNIST            & Resnet 44  & 0.66   & 43.5  & 45.2 & 45.5       & 0.38         & 72 & 86.59\\
FashionMNIST            & RepVGG-A0  & 7.84  & 54.0  & 54.6         & 55.3     & 0.52         & 99 & 88.88 \\ 
FashionMNIST            & RepVGG-A2  & 26.82   & 64.8  & 65.4 & 68.6     & 0.93        & 176 & 90.00\\
FashionMNIST           & ShufflenetV2-X1.5 (distilled)  & 2.48   & 52.0 & 53.0 & 55.3     & 0.40      & 76 & 87.76\\ 
% ImageNet             & Resnet 34 (distilled)       & 21.8   & 79.3          &  79.4       &   79.6       & 269 & 1.43  & 67.79  \\ 
\bottomrule
\end{tabular}
\caption {\textcolor{black}{Baseline measurements of performance and efficiency metrics on our selection of CNN models.}}
\label{table:singles}
\end{table*}

\begin{table*}[]

\centering
\begin{tabular}{@{}cllllcl@{}}
\toprule
\textbf{Identifier} &
  \multicolumn{1}{c}{\textbf{Description}} &
  \multicolumn{1}{c}{\textbf{First Model}} &
  \multicolumn{1}{c}{\textbf{Second Model}} &
  \multicolumn{1}{c}{\textbf{\begin{tabular}[c]{@{}c@{}}Score\\ Function\end{tabular}}} &
  \textbf{\begin{tabular}[c]{@{}c@{}}Threshold\\ Value\end{tabular}} &
  \multicolumn{1}{l}{\textbf{Memory}} \\ \midrule
CI1 & Single large model        & RepVGG-A2      & \multicolumn{1}{c}{-} & \multicolumn{1}{c}{-} & -      & -          \\
CI2 & Big/little representation & RepVGG-A0      & RepVGG-A2             & Difference            & 0.99   & -          \\
CI3 & Small complementary pair  & Resnet 20      & MobileNetV2-0.5       & Difference            & 0.8724 & No         \\
CI4 & Small complementary pair  & Resnet 20      & MobileNetV2-0.5       & Difference            & 0.8724 & DHash      \\
CI5 & Small complementary pair  & Resnet 20      & MobileNetV2-0.5       & Difference            & 0.8724 & Invariants \\
CI6 & Single large model (pruned)        & RepVGG-A2      & \multicolumn{1}{c}{-} & \multicolumn{1}{c}{-} & -      & -          \\
CI7 & Single small model (distilled)       & ShuffleNet V2 x1.5     & \multicolumn{1}{c}{-} & \multicolumn{1}{c}{-} & -      & -          \\

IM1 & Single large model        & RegNet-X-8GF   & \multicolumn{1}{c}{-} & \multicolumn{1}{c}{-} & -      & -          \\
IM2 & Big/little representation & RegNet-X-800MF & RegNet-X-8GF          & Difference            & 0.9074 & -          \\
IM3 & Small complementary pair  & MnasNet-1.3    & Densenet-121          & Difference            & 0.0983 & No         \\
IM4 & Small complementary pair  & MnasNet-1.3    & Densenet-121          & Difference            & 0.0983 & DHash      \\
IM5 & Small complementary pair  & MnasNet-1.3    & Densenet-121          & Difference            & 0.0983 & Invariants \\ 
IM6 & Single large model (pruned)        & RegNet-X-8GF   & \multicolumn{1}{c}{-} & \multicolumn{1}{c}{-} & -      & -          \\

IN1 & Single large model        & RepVGG-A2      & \multicolumn{1}{c}{-} & \multicolumn{1}{c}{-} & -      & -          \\
IN2 & Big/little representation & RepVGG-A0      & RepVGG-A2             & Difference            & 0.6  & -          \\
IN3 & Small complementary pair  & MobileNetV2-1.0  & Resnet 44    & Difference            & 0.2166 & No         \\
IN4 & Small complementary pair  & MobileNetV2-1.0   & Resnet 44    & Difference            & 0.2166 & DHash      \\
IN5 & Small complementary pair  & MobileNetV2-1.0   & Resnet 44     & Difference            & 0.2166 & Invariants \\
IN6 & Single large model (pruned)        & RepVGG-A2      & \multicolumn{1}{c}{-} & \multicolumn{1}{c}{-} & -      & -          \\
IN7 & Single small model (distilled)       & ShuffleNet V2 x1.5     & \multicolumn{1}{c}{-} & \multicolumn{1}{c}{-} & -      & -          \\

FM1 & Single large model        & RepVGG-A2      & \multicolumn{1}{c}{-} & \multicolumn{1}{c}{-} & -      & -          \\
FM2 & Big/little representation & RepVGG-A0      & RepVGG-A2             & Difference            & 0.9   & -          \\
FM3 & Small complementary pair  & Resnet 44 & MobileNetV2-1.0       & Max Probability            & 0.7917 & No         \\
FM4 & Small complementary pair  & Resnet 44 & MobileNetV2-1.0      & Max Probability             & 0.7917 & DHash      \\
FM5 & Small complementary pair  & Resnet 44 & MobileNetV2-1.0       & Max Probability             & 0.7917 & Invariants \\
FM6 & Single large model (pruned)        & RepVGG-A2      & \multicolumn{1}{c}{-} & \multicolumn{1}{c}{-} & -      & -          \\
FM7 & Single small model (distilled)       & ShuffleNet V2 x1.5     & \multicolumn{1}{c}{-} & \multicolumn{1}{c}{-} & -      & -          \\

% I3 & Single small model (distilled)        & Resnet 34   & \multicolumn{1}{c}{-} & \multicolumn{1}{c}{-} & -      & -          \\
\bottomrule
\end{tabular}
\caption {Experimental configurations.}
\label{table:configurations}
\end{table*}
% \color{black}

In the second approach, shown on the right side of Figure \ref{fig:fine-tuning}, we again select a complementary pair of models and pass the validation dataset through the first model as before. However, this time, we also pass the validation dataset through the second model, capturing the instances where predictions are correct. We then combine the incorrect predictions from the first model with the correct predictions from the second model into a single subset, which is used to train the second model. Next, we reverse the process by passing the validation dataset through both models again. This time, we collect the failed instances from the fine-tuned second model and the correct instances from the first model, combine them into a single subset, and use it to train the first model. This method aims to prevent the overfitting observed in the first approach, which occurs when training on a small subset of data.

% \subsection{Ablation with Score Functions}
%https://en.wikipedia.org/wiki/Ablation_(artificial_intelligence)

% \begin{figure}[ht!]
% \centering
% \includegraphics[width=0.5\textwidth]{Figures/cifar10-acc-plot--resnet20-mobilenetv2_x0_5-ablation.png}
% \caption {The use of the post-check mechanism achieves higher possible accuracy for the selected CNN pair. The vertical lines show the found threshold hyper parameter that achieves the best accuracy.}
% \label{fig:postcheck}
% \end{figure}

\section{Experimental Evaluation}
\label{Sec:ExpEval}
In this section, we present the experimental evaluation of our proposed methodology. Our objective is to assess the performance and efficiency of our method, utilizing Accuracy, Precision, Recall, and F1 Score for performance evaluation, as well as measuring energy consumption, current, and tail latency.

Subsections \ref{SubSec:ExpEval_Inf}, \ref{SubSec:ExpEval_Dat}, and \ref{SubSec:ExpEval_EvalMetrics} detail the experimental edge device, the datasets used, and the evaluation metrics, respectively. The CNN models used in our proposed methodology are described in subsection \ref{SubSec:ExpEval_CNNModels}. In subsection \ref{SubSec:ExpEval_Outcomes}, we summarize the experimental outcomes and discuss the evaluation results.% in subsection \ref{SubSec:ExpEval_Discussion}.

\subsection{Edge Device}
\label{SubSec:ExpEval_Inf}
% \cite{valladares2021performance}
We conducted our experiments on a Jetson Nano computer with a Quad-core ARM Cortex-A57 MPCore processor, 4GB RAM, and 128 NVIDIA CUDA cores, operating in 5W mode. The system OS was Ubuntu 20.04.6 LTS, and for the machine learning framework we used the PyTorch 1.13.0 library with Python 3.6. The experiments’ source code is available on GitHub \footnote{https://github.com/michaelkinnas/Reducing-Inference-Energy-Consumption-Using-Two-Complementary-CNNs}. The Jetson was powered through the USB at  5.15V, and we used a USB power meter capable of measuring milliamps (mAh) and watt-hours (Wh) to the second decimal digit.

\subsection{Datasets} 
\label{SubSec:ExpEval_Dat}

% We conducted our experiments on CIFAR-10 \cite{krizhevsky_learning_2012}, comprising 10,000 validation images of size 3x32x32 spread across 10 classes, and ImageNet \cite{deng_imagenet_2009}, comprising 50,000 images of size 3x224x224 spread across 1000 classes.
% More specifically, the ImageNet model architectures are designed to receive 3x224x224 size inputs, but the actual images in the ImageNet dataset have higher varying widths and heights. The vast majority of these images contain 3 channels (RGB), while a few are grayscale with only 1 channel. The Jetson Nano was unable to load all 50,000 test images into main memory, at their default dimensions, as this would require significantly more than the 4GB of available RAM. To address this limitation, we selected a subsample of 10,000 images from the test dataset, comprising of 10 images from each of the 1,000 classes. We pre-processed these images by resizing them to 224x224 pixels and converting grayscale images to 3-channel images.

We conducted our experiments on CIFAR-10 \cite{krizhevsky_learning_2012}, comprising 10,000 validation images of size 3x32x32 spread across 10 classes, Intel Image Classification \footnote{https://www.kaggle.com/datasets/puneet6060/intel-image-classification}, comprising of 3,000 validation images of size 3x150x150 spread across 6 classes, FashionMNIST \cite{xiao2017fashion}, comprising of 10,000 images of size 1x28x28 spread across 10 classes and ImageNet \cite{deng_imagenet_2009}, comprising 50,000 images of size 3x224x224 spread across 1,000 classes.
More specifically, the ImageNet model architectures are designed to receive 3x224x224 size inputs, but the actual images in the ImageNet dataset have higher varying widths and heights. The vast majority of these images contain 3 channels (RGB), while a few are grayscale with only 1 channel. The Jetson Nano was unable to load all 50,000 test images into main memory, at their default dimensions, as this would require significantly more than the 4GB of available RAM. To address this limitation, we selected a subsample of 10,000 images from the test dataset, comprising of 10 images from each of the 1,000 classes. We pre-processed these images by resizing them to 224x224 pixels and converting grayscale images to 3-channel images.

\begin{figure}[ht!]
\centering
\includegraphics[width=0.5\textwidth]{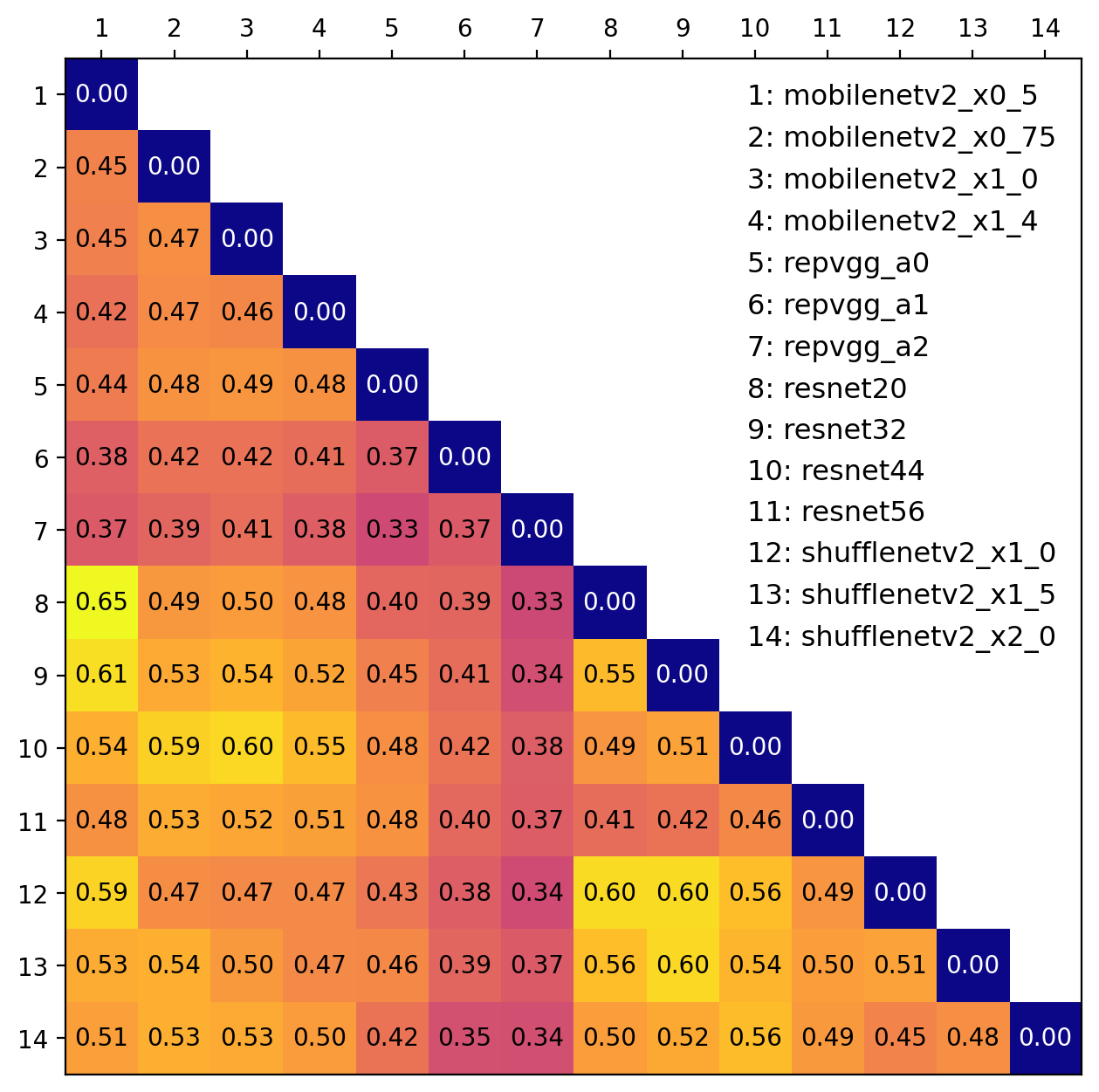}
\caption {Complementarity matrix of CIFAR-10 available pretrained PyTorch models.}
\label{fig:het-matrix}
\end{figure}

To determine the optimal hyperparameter $\lambda^*$ for achieving the highest accuracy, as mentioned in equation (\ref{find-best-lambda}), we can use a validation dataset or a part of the training dataset. For ImageNet, we utilized the additional 40,000 images from the test set as our validation dataset. However, for the CIFAR-10 dataset, which does not have a designated validation dataset, we employed the training set for this purpose. 
To test the memory component we duplicated our samples for each dataset at different ratios, while also applying simple transformations such as rotations and mirroring on the duplicates, and run the same experiments once using the difference hash method and once using the invariants from complex moment method as described in subsection \ref{SubSec:ProposedMethodology_Mem}.

\subsection{Evaluation Metrics}
\label{SubSec:ExpEval_EvalMetrics}
%For our evaluation metrics, we utilized various indicators of performance and efficiency.
We measured performance using four metrics: Accuracy, Precision, Recall, and F1-score. 
Accuracy is defined as the ratio of correctly predicted instances to the total instances. Precision is the ratio of correctly predicted positive instances to the total predicted positives. Recall is the ratio of correctly predicted positive instances to all actual positives. F1-score is the harmonic mean of Precision and Recall, providing a single metric that balances both concerns. Precision, Recall, and F1-score are calculated from the perspective of a specific class, but for our multi-class classification task, we averaged these metrics across all classes.
For energy consumption, we measured the current in milliamps (mAh) and energy consumption in watt hours (Wh) used throughout the entire inference process.

% \begin{figure}[ht!]
% \centering
% \includegraphics[width=0.5\textwidth]{Figures/mnasnet1_3 - densenet121 acc curves diff 2.png}
% \caption {Post-check Evaluation}
% \label{fig:postcheck}
% \end{figure}

\begin{figure}[ht!]
\centering
\includegraphics[width=0.5\textwidth]{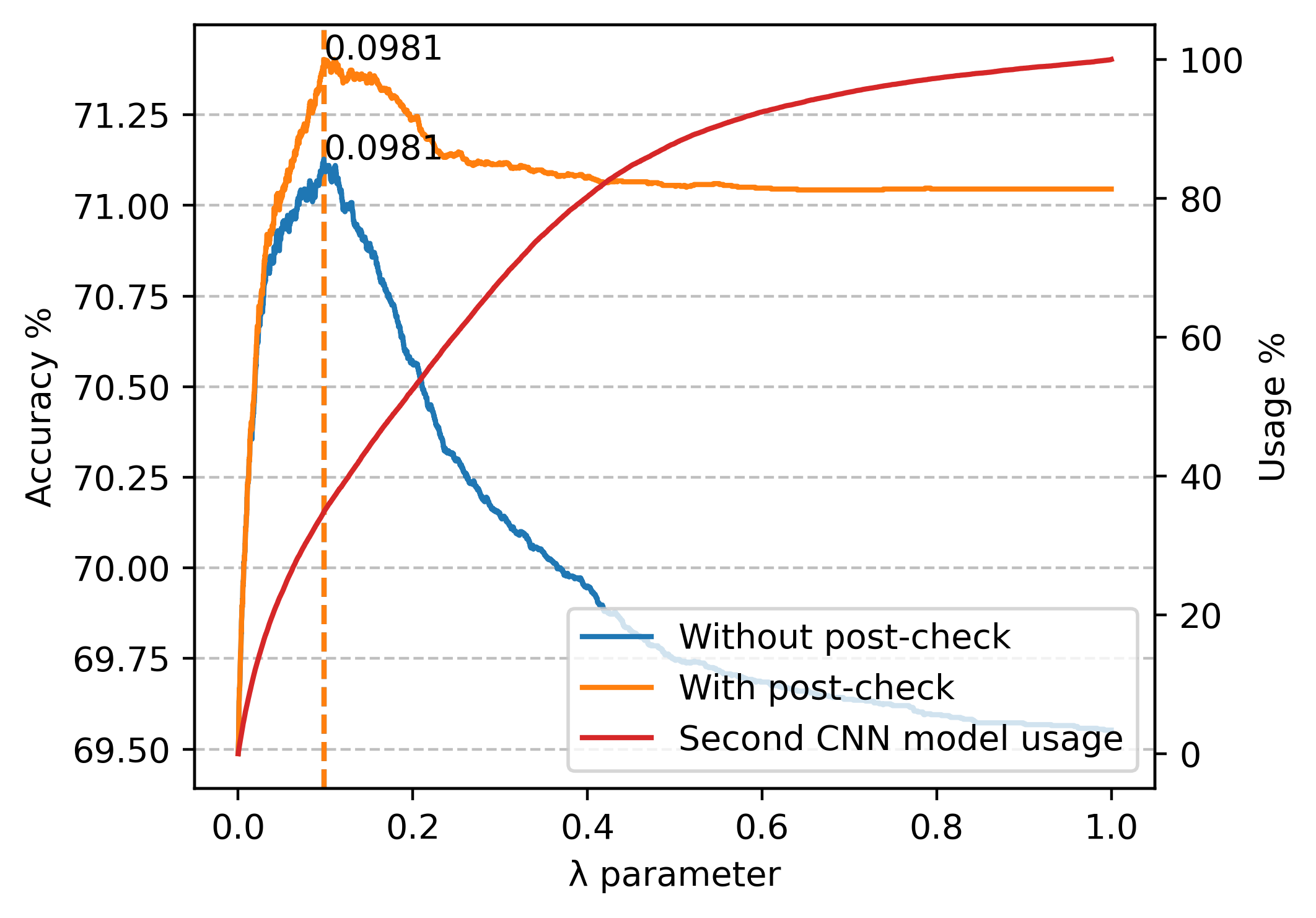}
\caption {Post-check Evaluation, using the Difference score function on the ImageNet validation dataset using our configuration I3}
\label{fig:postcheck}
\end{figure}

% \begin{figure*}[ht!]
% \centering
% \includegraphics[width=\textwidth]{Figures/energy.png}
% \caption {Current and Energy consumption measurements of tested configurations for (a) CIFAR-10 and (b) ImageNet datasets.}
% \label{fig:energy}
% \end{figure*}

%Major review
\begin{figure*}[ht!]
\centering
\includegraphics[width=\textwidth]{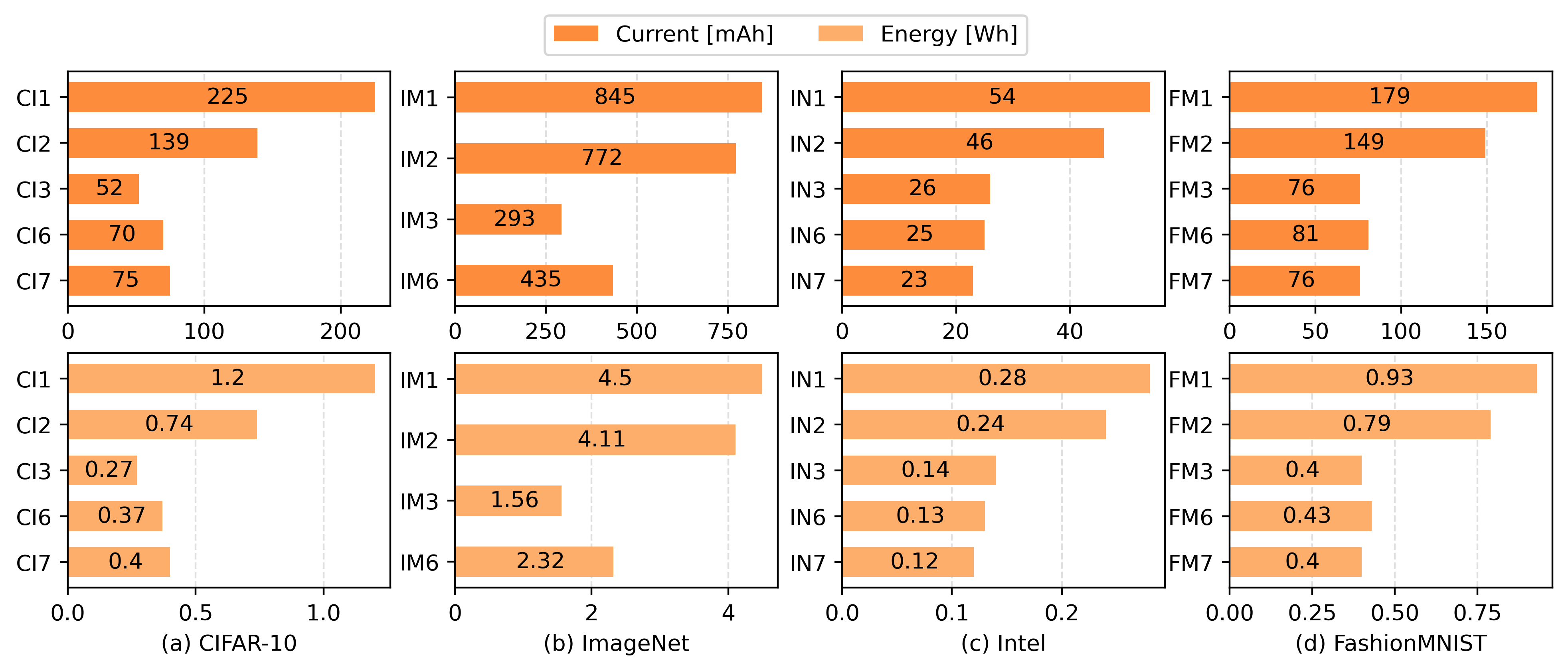}
\caption {Current and Energy consumption measurements of tested configurations for (a) CIFAR-10, (b) ImageNet, (c) Intel and (d) FashionMNIST datasets.}
\label{fig:energy}
\end{figure*}

We conducted the inference process for each dataset, processing one sample at a time to measure response times. We recorded the mean response time in milliseconds (ms), as well as the 95-th and 99-th percentile tail latencies, defined as the maximum response times that 95\% and 99\% of the input samples respectively experienced from the time a sample enters the system to when it produces a result. Tail latency \cite{theodoropoulos_new_2024} is a more important evaluation metric than mean response time in latency-sensitive edge computing systems. The reason is that tail latency directly impacts user satisfaction and service level agreements by ensuring that even a small percentage of users do not face unacceptably long delays. It provides a more comprehensive evaluation metric of on-device artificial intelligence applications by highlighting extreme values in the response time distribution, which traditional metrics may overlook.%, thus offering a more accurate representation of the user experience.

\subsection{CNN Models}
\label{SubSec:ExpEval_CNNModels}
We utilized benchmark CNN models available from PyTorch Hub \footnote{https://pytorch.org/hub/}. Our selection criteria ensured that the largest model could run on the Jetson Nano device while achieving high prediction accuracy, along with its smaller variant.
% These two models will serve as the big/little implementation for our comparisons. The large CNN also represents the single large model used in our experimental comparison.
Two models from PyTorch Hub will be used as the basis for our big/little implementation in the comparisons. The large CNN also serves as the single large model in our experimental analysis. Additionally, we included a pruned version of this large model, along with a smaller model trained through knowledge distillation from the large model. This allows us to compare our methodology with established model compression techniques for a comprehensive evaluation. More specifically, with the pruned model, we began with the large baseline model and systematically pruned it to a size comparable to the combined size of the CNN pair used in our methodology, ensuring fair comparison. For the small distilled model, we started with a single, similarly-sized small model and applied a standard knowledge distillation method, using the large model as the teacher. It is important to note that we do not include a distilled model for the ImageNet dataset due to the substantial challenges in training with this large dataset, particularly in terms of time and computational resources. However, we have provided a distilled model for the other three datasets, CIFAR-10, Intel and FashionMNIST.

We did not find any comprehensive list of pretrained models for the Intel and FashionMNIST datasets, therefore, we used the already pretrained models from CIFAR-10 and applied standard transfer learning, fine-tuning all layers for a few epochs.

To identify suitable pairs of CNNs for our proposed methodology, we applied the complementarity formula (\ref{complementarity}) to the models available in PyTorch Hub, resulting in the complementarity matrix as shown in Figure \ref{fig:het-matrix}. This matrix displays all possible combinations of model pairs and their corresponding complementarity scores for the CIFAR-10 dataset. The highest score was observed between ResNet-20 and MobileNetV2-0.5, making this pair our selection for our methodology. % configurations. 
For better illustration, all values are multiplied by 10.
The diagonal elements of the matrix have a complementarity score of 0, as identical models produce identical predictions, representing full homogeneity.
%The diagonal elements of the matrix, where each model is compared against itself, yield a complementarity score of 0. This occurs because identical models produce identical predictions, resulting in 100\% intersection and thus representing a fully homogeneous pair. 

We employed a similar process for selecting the model pair for the ImageNet dataset which we do not illustrate for the sake of brevity.
After selecting the most suitable models, we evaluated each one individually to establish a baseline. The results are summarized in Table \ref{table:singles}.

% \begin{figure*}[ht!]
% \centering
% \includegraphics[width=\textwidth]{Figures/performance_2.png}
% \caption {Performance of tested configurations for (a) CIFAR-10 and (b) ImageNet datasets.}
% \label{fig:performance}
% \end{figure*}

%Major revision
\begin{figure*}[ht!]
\centering
\includegraphics[width=\textwidth]{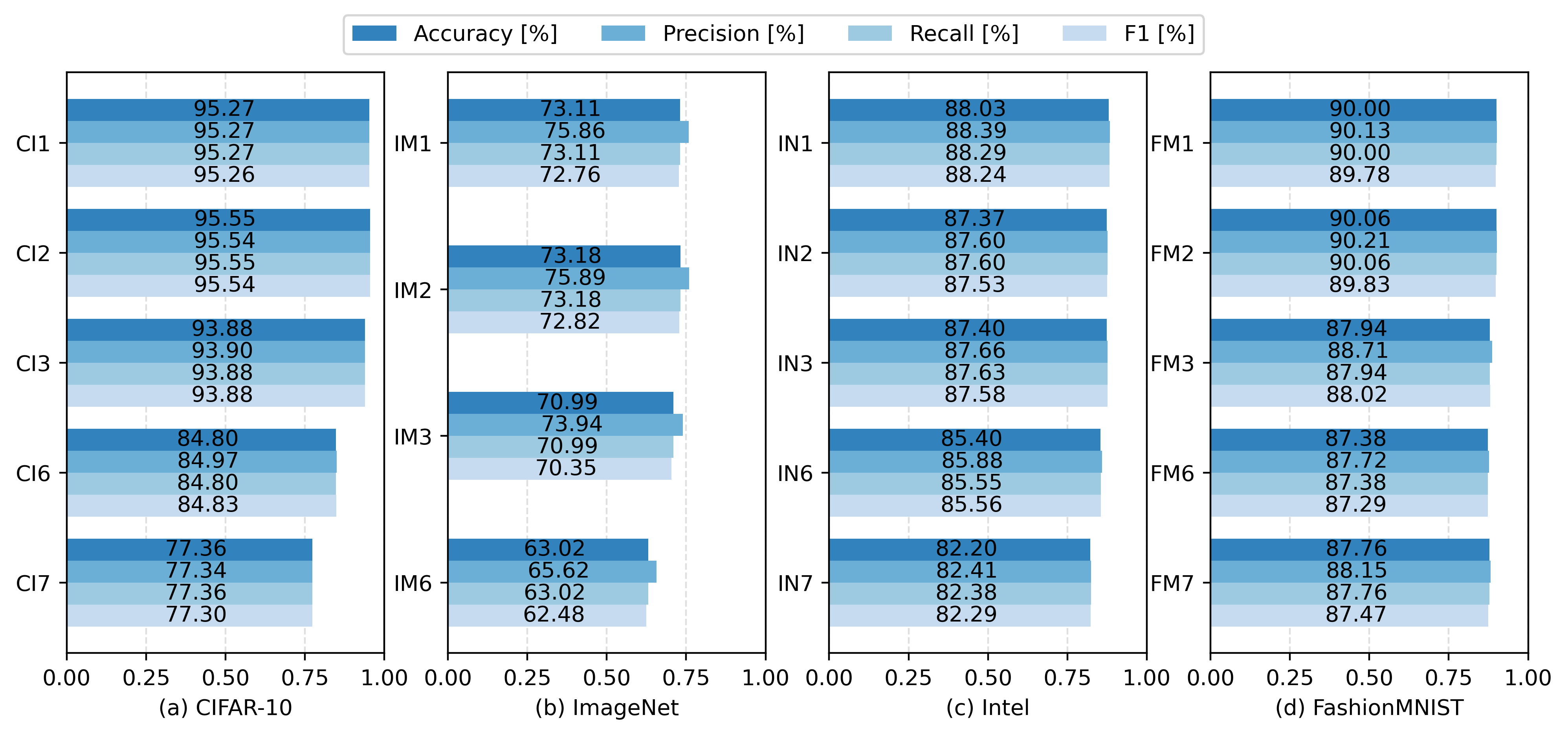}
\caption {\textcolor{blue}{Performance of tested configurations for (a) CIFAR-10, (b) ImageNet, (c) Intel and (d) FashionMNIST datasets.}}
\label{fig:performance}
\end{figure*}

% Table \ref{table:configurations} details all the experimental configurations discussed in this paper including a single large model, a pair of a large and a small model and three pairs of small complementary models for every dataset. Each configuration is identified by a prefix letter representing the dataset (C for CIFAR-10 and I for ImageNet) and a suffix number indicating its sequence in the table. Our proposed methodology configurations are denoted by identifiers C3 to C5 and I3 to I5, while the configurations used for comparison are identified as C1 and I1 for the single large model, C2 and I2 for the big/little representation, C6 and I6 for the single large pruned model and C7 for the small distilled model. We will use this terminology in the subsections \ref{SubSec:ExpEval_Outcomes} of outcomes and discussions.

Table \ref{table:configurations} details all the experimental configurations discussed in this paper including a single large model, a pair of a large and a small model and four pairs of small complementary models for every dataset. Each configuration is identified by a prefix letter representing the dataset (CI for CIFAR-10, IM for ImageNet, IN for Intel and FM for FashionMNIST) and a suffix number indicating its sequence in the table. Our proposed methodology configurations are denoted by identifiers CI3 to CI5, IM3 to IM5 and IN3 to IN5 and FM3 to FM5 while the configurations used for comparison are identified as CI1,  IM1, IN1 and FM1 for the single large model, CI2, IM2, IN2 and FM2 for the big/little representation, CI6, IM6, IN6 and FM6 for the single large pruned model and CI7, IN7 and FM7 for the small distilled model. We will use this terminology in the subsections 
 \ref{SubSec:ExpEval_Outcomes} of outcomes and discussions.

Confidence score functions and threshold values for all experimental configurations using two CNNs were determined by applying the formula (\ref{find-best-lambda}). The final combination of score function and threshold value was selected based on the maximum achieved accuracy for each configuration.

%NEW Fine-Tuning
\subsection{Fine-tuning CNN pairs}
\label{SubSec:Fine-TuningCNNPairs}

To implement the fine-tuning methods, we first selected five pairs from the ImageNet dataset, with the highest complementarity using the complementarity matrix. We then applied the fine-tuning approaches outlined in Section \ref{SubSec:IncreasingComplementarity}. Once the fine-tuning was completed, we evaluated the fine-tuned pairs using our primary methodology as described in the rest of Section \ref{Sec:ProposedMethodology}. Finally, we compared the accuracy of the fine-tuned pairs against the same models with their default pre-trained weights obtained from PyTorch Hub.

\subsection{Outcomes } 
\label{SubSec:ExpEval_Outcomes}

% \begin{figure*}[ht!]
% \centering
% \includegraphics[width=\textwidth]{Figures/latency.png}
% \caption {Response times of tested configurations for (a) CIFAR-10 and (b) ImageNet datasets.}
% \label{fig:response}
% \end{figure*}

\begin{figure*}[ht!]
\centering
\includegraphics[width=\textwidth]{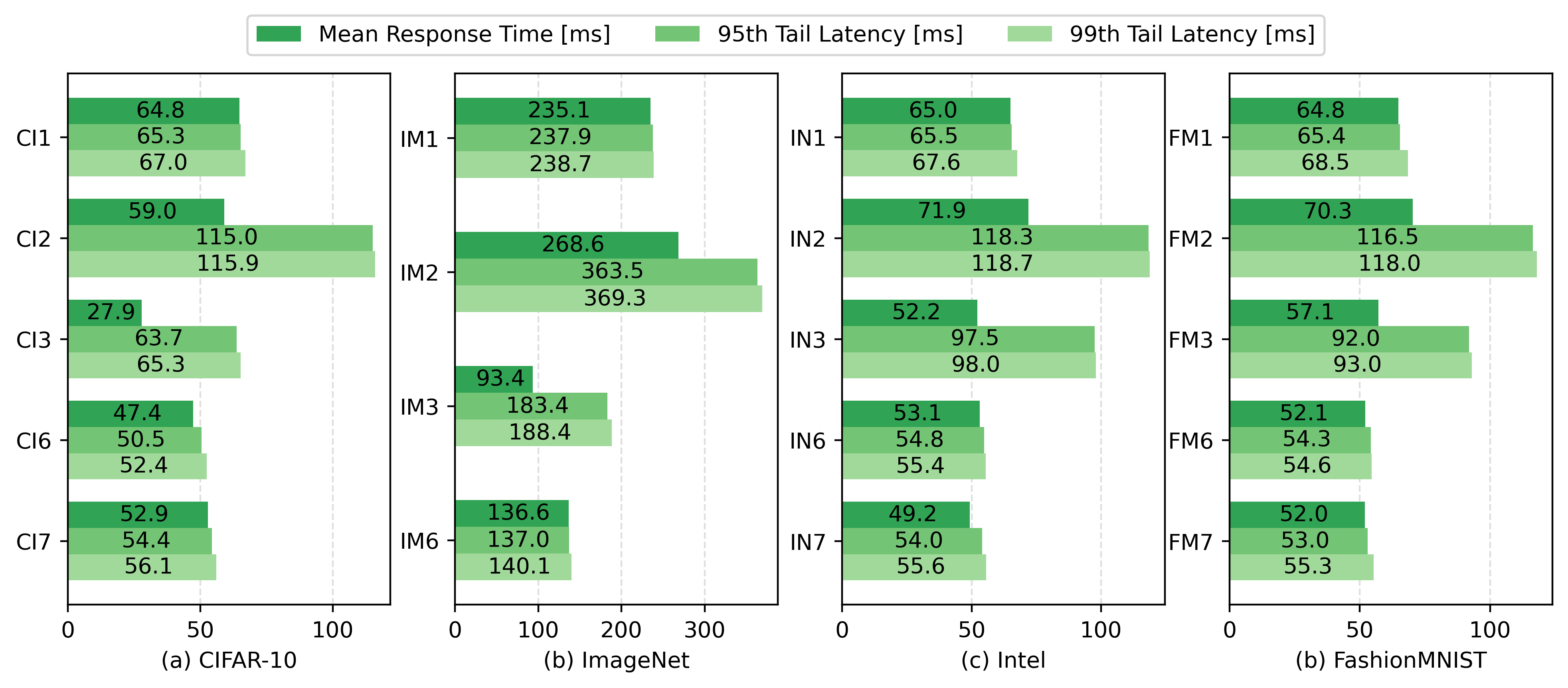}
\caption {Response times of tested configurations for (a) CIFAR-10, (b) ImageNet, (c) Intel and (d) FashionMNIST datasets.}
\label{fig:response}
\end{figure*}

\subsubsection{Post-check Ablation}
First, we want to examine the application of the post-check mechanism, described in subsection \ref{SubSec:ProposedMethodology_PostCheck}. 
The Figure \ref{fig:postcheck} shows the accuracy in the validation dataset (left axis) with post-check (orange line) and without post-check (blue line) in the IM3 configuration which includes the models MnasNet-1.3 and Densenet-121 using the ImageNet dataset.
The vertical lines show the selected threshold hyperparameter $\lambda^*$, 0.0983 in this case,  that achieves the best accuracy, for this specific configuration.
By incorporating post-check, we observe a slight enhancement in the total accuracy of the setup. %The use of the post-check mechanism achieves higher possible accuracy for the selected CNN pair. 
In addition we see the percentage of usage (right axis) of the second model (red line) as a function of the $\lambda$ hyperparameter. As the value of $\lambda$ increases, the usage of the second model also increases.

% Πόσο τις εκατό χρησιμοποιείται το δεύτερο μοντέλο. Επηρεάζεται από την τιμή του λ. Όσο πιο μεγάλη τόσο πιο αυστηρός ο έλεγχος και τόσο πιο πολύ θα χρησιμοποιηθεί το δεύτερο CNN.
%Η μπλε και η πορτοκαλή γραμμή (without post-check και with post-check αντίστοιχα) είναι το accuracy curve σε συνάρτηση με την τιμή υπερπαραμέτρο λ. Η κόκκινη γραμμή (second model usage) είναι το ποσοστό χρήσης του δεύτερου μοντέλου σε συνάρτηση με την τιμή της υπερπαραμέτρου λ. Στην ουσία είναι δύο γραφήματα σε ένα. Γι αυτό υπάρχουν δύο άξονες y.

%Major revision

\subsubsection{$\lambda$ Sensitivity Analysis}

\begin{figure}[ht!]
\centering
\includegraphics[width=0.5\textwidth]{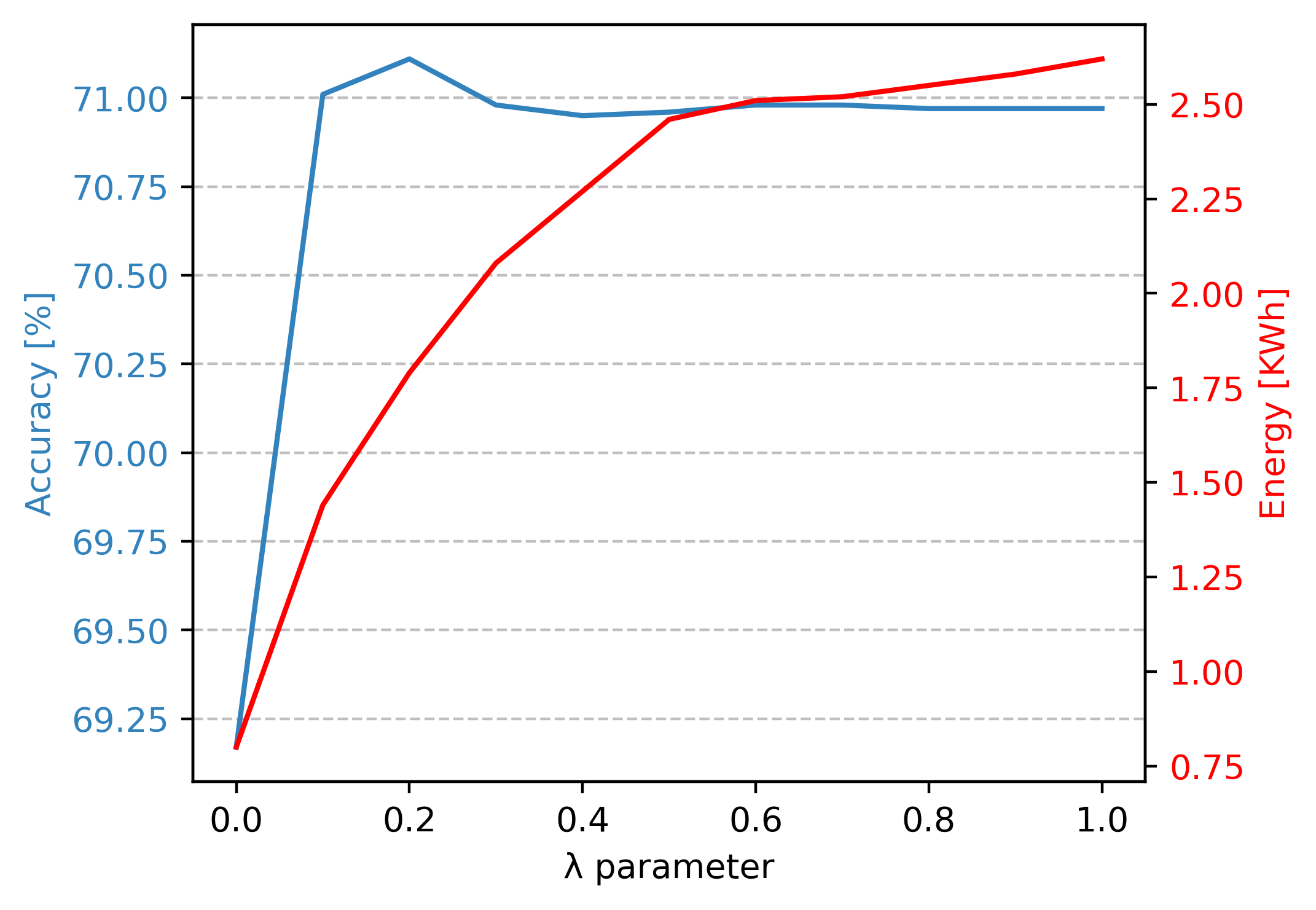}
\caption {Accuracy and energy curves as a function of $\lambda$ parameter, for IM3 configuration on ImageNet test set.}
\label{fig:sensitivity}
\end{figure}

We examine the relationship between accuracy and energy consumption as a function of the $\lambda$ parameter. 
As shown in Figure \ref{fig:sensitivity}, increasing the $\lambda$ parameter tightens the acceptance threshold for predictions from the first model. As a result, the second model is invoked more frequently, leading to higher energy consumption. In this figure, the blue curve represents the accuracy as $\lambda$ varies, with its corresponding y-axis on the left, while the red curve depicts the total energy consumed during inference on the test set, with its y-axis on the right. When $\lambda$ approaches 0, the first model predominantly generates the predictions, resulting in lower overall accuracy, similar to the performance of using only the first model, as only its predictions are accepted. As $\lambda$ increases, the second model is invoked more frequently. Additionally, the figure shows that accuracy remains largely stable, except for very small values of $\lambda$. This stability is maintained by the Post-check mechanism, which ensures that the most reliable prediction is selected for each input. Without this mechanism, as shown by the blue curve in Figure \ref{fig:postcheck}, accuracy declines because the second model's predictions are accepted unconditionally when the confidence threshold of the first model is not met.
\color{black}

\subsubsection{Score Functions Comparison}

% major review v1
% To evaluate the proposed score functions, we conducted a threshold hyperparameter search (\ref{eq:acc}) across all three score functions and tested our methodology using the standard ImageNet configuration (I5) on the test set. Figure \ref{fig:scorefn_comparisson} presents the accuracy results for each score function with its corresponding optimal $\lambda^*$ value. From the figure, it is evident that, in this specific configuration, the Difference score function achieves the highest performance.

\begin{figure}[ht!]
\centering
\includegraphics[width=0.5\textwidth]{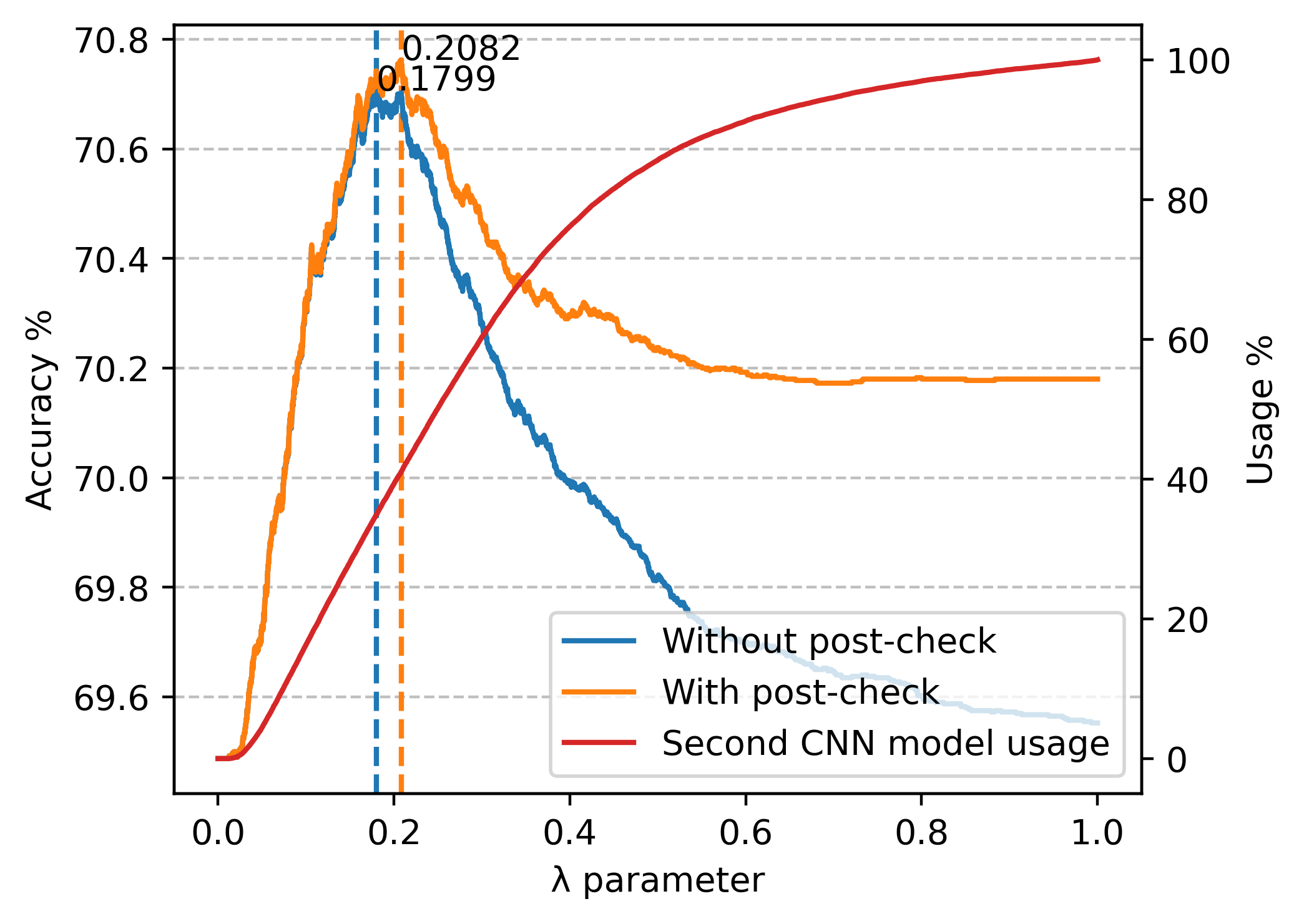}
\caption {Post-check evaluation for the Max Probability score function on the ImageNet validation dataset using configuration \textcolor{blue}{IM3}.}
\label{fig:post-check-maxp}
\end{figure}

\begin{figure}[ht!]
\centering
\includegraphics[width=0.5\textwidth]{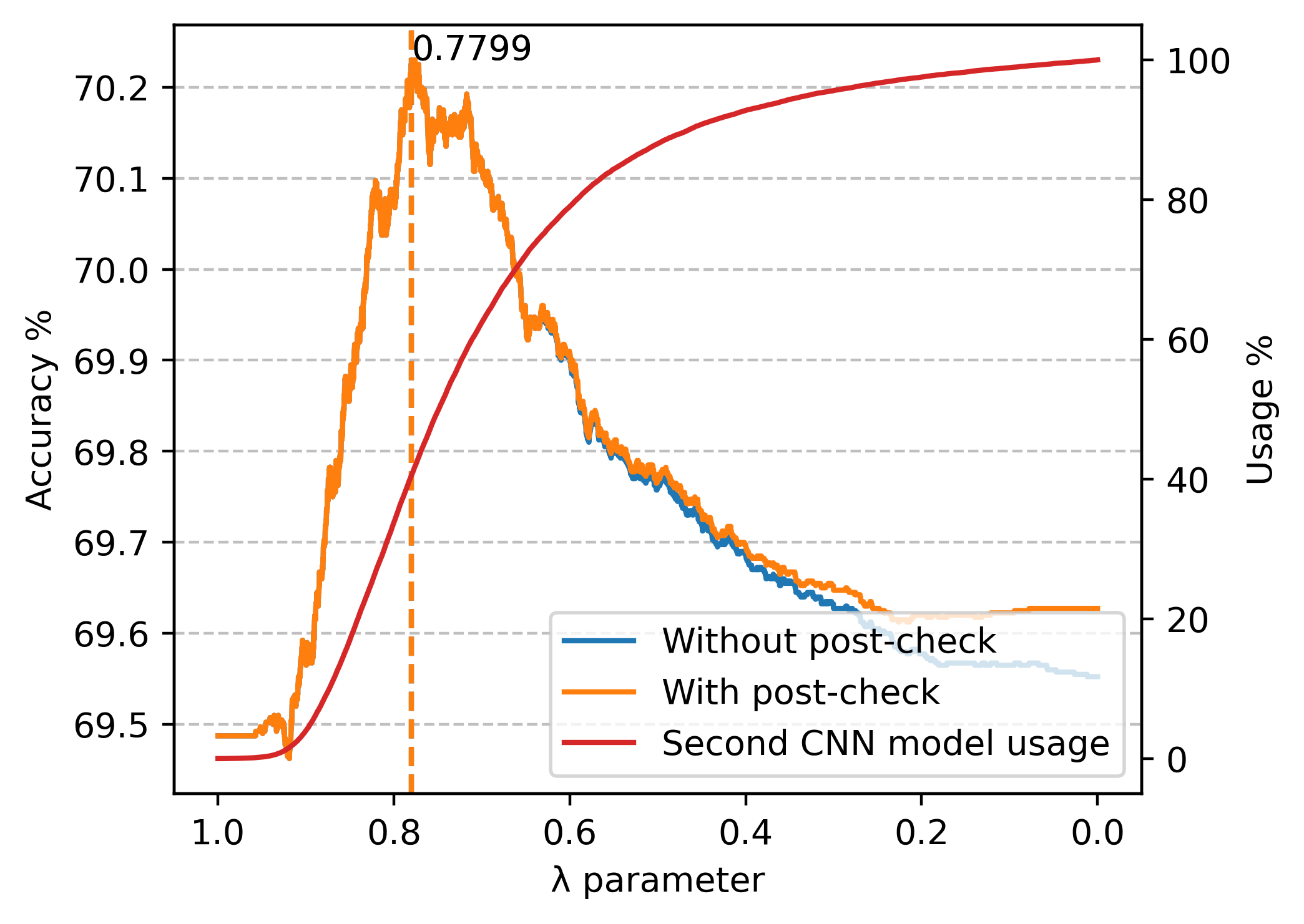}
\caption {Post-check evaluation for the Entropy score function on the ImageNet validation dataset using configuration \textcolor{blue}{IM3}.}
\label{fig:post-check-entropy}
\end{figure}

% major review v2
To gain a deeper understanding of the selection process among the three available score functions, we present a comparative performance analysis using the $\lambda^*$ parameter search function (Equation \ref{find-best-lambda}). Two additional figures are included to illustrate the Post-check evaluation for the Max Probability score function (Equation \ref{max_P}) in Figure \ref{fig:post-check-maxp} and for the Entropy score function (Equation \ref{entropy-norm}) in Figure \ref{fig:post-check-entropy}. These are compared alongside the previously discussed Difference score function (Equation \ref{difference}) shown in Figure \ref{fig:postcheck}.

From the comparison of these figures, it is evident that for our specific ImageNet configuration (IM3), utilizing the pair of CNNs MnasNet-1.3 and Densenet-121, the best performance is achieved with the Difference score function at $\lambda^* = 0.0981$, yielding the highest accuracy on the validation dataset. Notably, the Entropy score function performs the worst for this particular CNN pair. Additionally, it is observed that in all three cases, the use of the Post-check mechanism consistently results in either improved or, at minimum, comparable performance across the full range of $\lambda$ values.

\color{black}

% \begin{figure}[ht!]
% \centering
% \includegraphics[width=0.5\textwidth]{Figures/compare_score_fns.png}
% \caption {\textcolor{blue}{Accuracy performance using the three proposed score functions with their corresponding $\lambda^*$ parameter values, with the I5 configuration on the ImageNet test set.}}
% \label{fig:scorefn_comparisson}
% \end{figure}

\subsubsection{Energy Evaluation}

Figure \ref{fig:energy} presents a comparison of energy consumption across different experimental configurations (Table \ref{table:configurations}) for the CIFAR-10, ImageNet, Intel and FashionMNIST datasets. For the CIFAR-10 dataset, our configuration (CI3) demonstrates a substantial reduction in energy consumption, achieving a 62.6\% decrease compared to the big/little configuration (CI2) and a 76.9\% decrease relative to the single large model (CI1). Additionally, it reduces energy consumption by 25.6\% compared to the large pruned model (CI6) and by 30.7\% when compared to the small distilled model (CI7).

A similar trend is observed with the ImageNet dataset. Our configuration (IM3) reduces energy consumption by 62\% relative to the big/little configuration (IM2) and by 65.3\% when compared to the single large model (IM1). In comparison to the large pruned model (IM6), energy consumption is decreased by 32.6\%.

For the Intel dataset, our configuration (IN3) achieves a 43.5\% reduction in energy consumption compared to the big/little configuration (IN2) and a 52.9\% reduction relative to the single large model (IN1). However, it shows a slight 4\% increase in energy consumption compared to the large pruned model (IN6) and a 13\% increase compared to the small distilled model (IN7).

Lastly, for the FashionMNIST dataset, our configuration (FM3) results in a 49\% decrease in energy consumption compared to the big/little configuration (FM4) and a 57.5\% reduction relative to the single large model (FM1). It also achieves a 6.2\% reduction in energy consumption compared to the large pruned model (FM6), while consuming an equivalent amount of energy as the small distilled model (FM7).
\color{black}

\begin{figure}[ht!]
\centering
\includegraphics[width=0.5\textwidth]{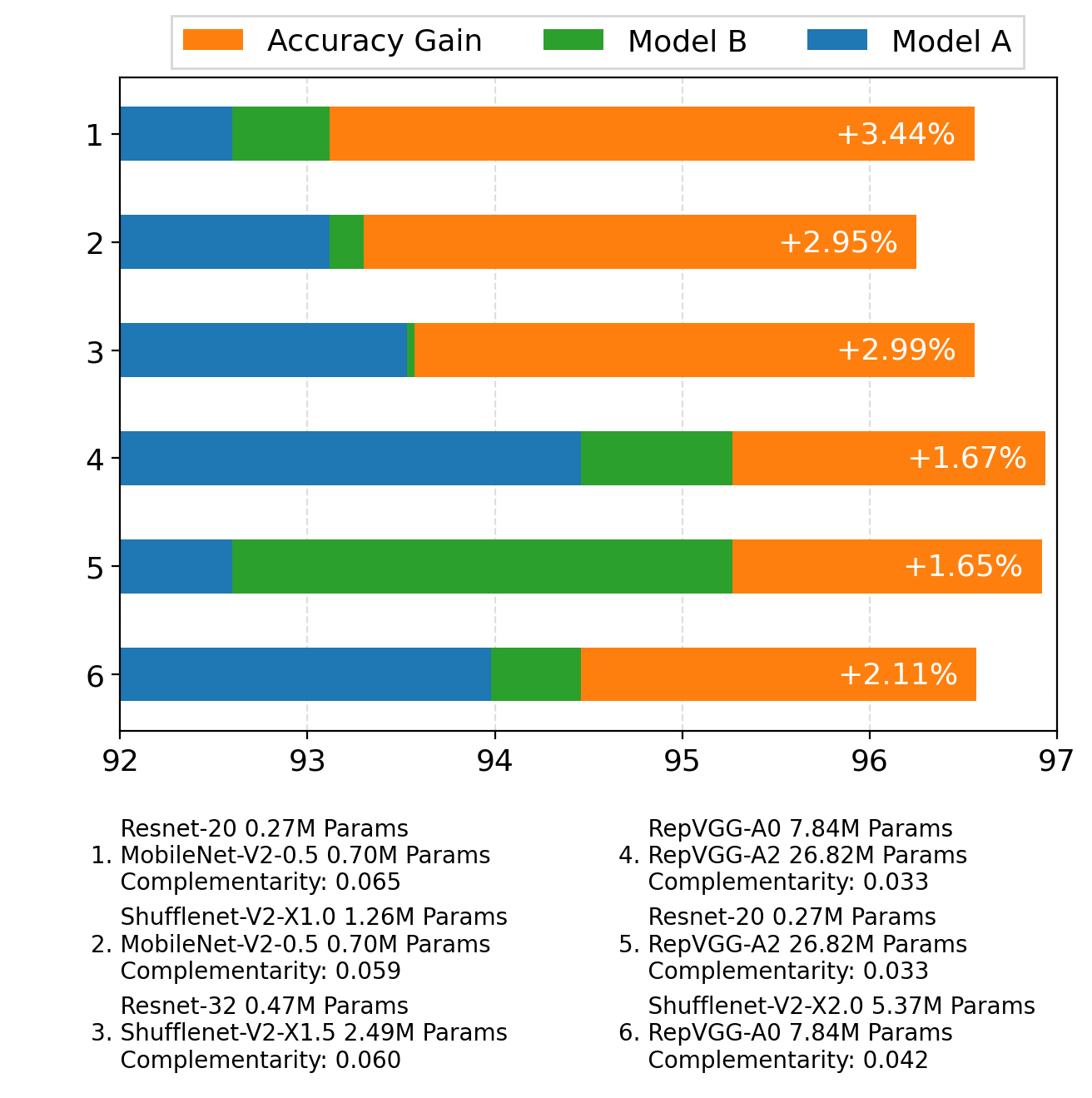}
\caption {Accuracy gains as a function of the complementarity of two CNN pairs.}
\label{fig:coverage-gain}
\end{figure}

To evaluate the system's performance in a dynamic IoT setting with constantly changing inputs, we conducted an experiment. We provided 10 randomly selected batches (each containing 100 images) from CIFAR-10 and 10 batches (each containing 100 images) from ImageNet, which were fed to the Jetson Nano device in a random order. The Jetson Nano forwarded each image to the DNNs presented in Table \ref{table:configurations}. This setup created a highly dynamic scenario with continuously changing and unpredictable input streams. Power consumption was measured using a power meter, and the results showed that the approach with the Single large models consumed 0.57 Wh, the Big/Little models consumed 0.484 Wh, and the small complementary pairs consumed 0.177 Wh. These findings demonstrate that the proposed approach effectively balances energy efficiency in high-throughput, variable data streams and outperforms other methods in this dynamic environment.

\subsubsection{Performance Evaluation}
Figure \ref{fig:performance} illustrates the inference performance metrics.
% For the CIFAR-10 dataset, our configuration (\textcolor{blue}{C5}) has 1.67\% performance degradation compared to the big/little configuration (\textcolor{blue}{C4}) and 1.39\% performance degradation compared to the single large model (C1). A similar trend is observed with the ImageNet dataset: our configuration (\textcolor{blue}{I5}) has 2.19\% performance degradation compared to the big/little setup (\textcolor{blue}{I4}) and  2.12\% performance degradation compared to the single large model (I1). 
% In addition we should clarify that classification in ImageNet generally has lower performance than in CIFAR-10 due to the significantly larger and more diverse set of images and classes in ImageNet, which increases the complexity and difficulty of the classification task. Although our methodology results in a slight decrease in performance metrics close to 2\% due to the use of significantly smaller models, it achieves substantial improvements in energy consumption close to 70\%, which is the primary objective of this research.
% Major revision

For the CIFAR-10 dataset, our configuration (CI3) shows a 1.67\% performance degradation compared to the big/little configuration (CI2) and a 1.39\% degradation relative to the single large model (CI1). However, when compared to the large pruned model (CI6), our configuration shows a 9.0\% improvement in performance, and a 16.5\% increase when compared to the small distilled model (CI7).

A similar trend is observed with the ImageNet dataset. Our configuration (IM3) shows a 2.19\% performance degradation relative to the big/little setup (IM2) and a 2.12\% degradation compared to the single large model (IM1). In contrast, our configuration demonstrates an 8.0\% performance improvement over the large pruned model (IM6).

For the Intel dataset, our configuration (IN3) shows a slight performance increase of 0.03\% compared to the big/little configuration (IN2), though there is a 0.57\% decrease in performance relative to the single large model (IN1). However, it achieves a 2.0\% performance improvement over the large pruned model (IN6) and a 3.2\% improvement compared to the small distilled model (IN7).

Lastly, for the FashionMNIST dataset, our configuration (FM1) exhibits a 2.12\% decrease in performance relative to the big/little configuration (FM2) and a 2.06\% decrease compared to the single large model (FM1). In contrast, it demonstrates a 0.56\% performance improvement over the large pruned model (FM6) and a 0.18\% increase relative to the small distilled model (FM7).

\color{black}

In addition we should clarify that classification in ImageNet generally has lower performance than in CIFAR-10, Intel and FashionMNIST due to the significantly larger and more diverse set of images and classes in ImageNet, which increases the complexity and difficulty of the classification task. Although our methodology results in a slight decrease in performance metrics close to 2\% due to the use of significantly smaller models, it achieves substantial improvements in energy consumption close to 70\%, which is the primary objective of this research.

%The accuracy gain by levering relatively higher complementary pairs (top 3) is higher than the individual same models compared to relatively lower complementary pairs (lower 3).

%To evaluate the improvement in accuracy achieved by leveraging the complementarity of the CNN pairs, we compare the performance of the pair configurations illustrated in Figure \ref{fig:performance} against the baseline performance of each individual model within each pair, as shown in Table \ref{table:singles}. For the CIFAR-10 dataset, our implementation (C3) exhibits a 0.76\% increase in prediction accuracy over the best-performing individual model it includes (i.e. MobileNetV2-0.5), while the big/little configuration (C4) shows a 0.28\% improvement against the single big CNNN (i.e. RepVGG-A2). Similarly, for the ImageNet dataset, our implementation (I3) achieves a 1.34\% increase in prediction accuracy compared to the highest-performing single model (i.e. DenseNet-121), whereas the big/little configuration (I2) demonstrates only a 0.07\% enhancement compared to the single big CNN (RegNet-X-8GF).

\begin{figure*}[ht!]
\centering
\includegraphics[width=\textwidth]{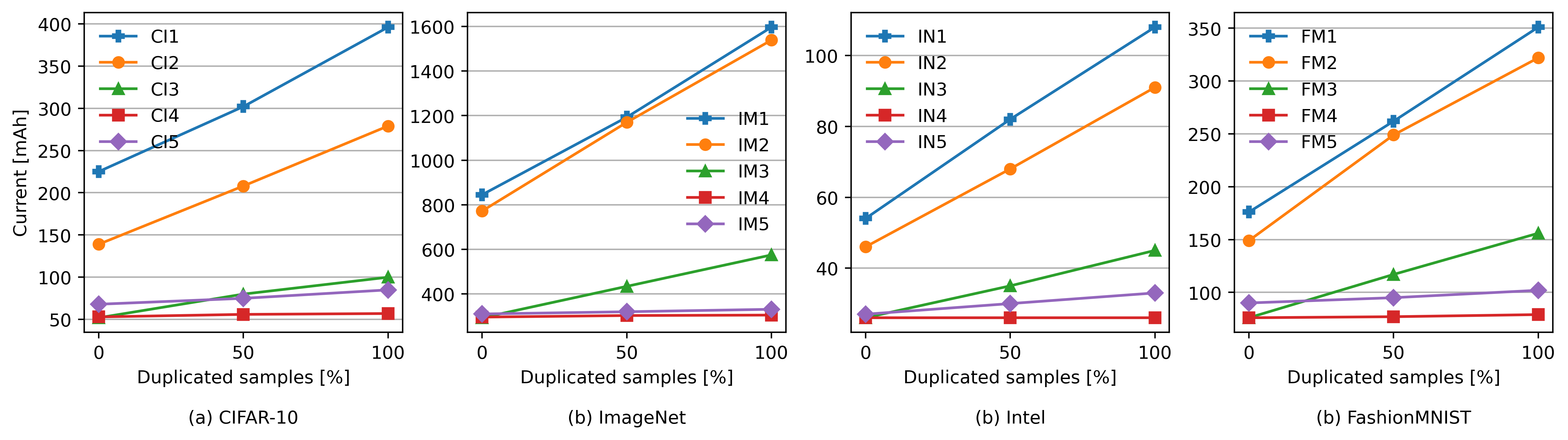}
\caption {Energy consumption comparisons using memory component with duplicated samples.}
\label{fig:memory}
\end{figure*}

\subsubsection{Complementarity Evaluation}
To evaluate the improvement in accuracy achieved by leveraging the complementarity of the CNN pairs we made experimental comparisons of the CNNs pairs against the single CNNs they include and six CNNs pairs with different complementarity values.
In Figure \ref{fig:performance} we see the performance of the pair configurations against the baseline performance of each individual model within each pair, as shown in Table \ref{table:singles}. 

For the CIFAR-10 dataset, our implementation (CI3) exhibits a 0.76\% increase in prediction accuracy over the best-performing individual model it includes (i.e. MobileNetV2-0.5), while the big/little configuration (CI2) shows a 0.28\% improvement against the single big CNN (i.e. RepVGG-A2). 

Similarly, for the ImageNet dataset, our implementation (IM3) achieves a 1.34\% increase in prediction accuracy compared to the highest-performing single model (i.e. DenseNet-121), whereas the big/little configuration (IM2) demonstrates only a 0.07\% enhancement compared to the single large CNN (RegNet-X-8GF). 

For the Intel dataset our configuration (IN2) achieves a 0.7\% increase in prediction accuracy over the best-performing individual model in includes (i.e. MobileNetV2-X1.0), while the big/little configuration (IN3) shows a 0.9\% decrease in predictive performance against the single big CNN (i.e. RepVGG-A2). 

Lastly for the FashionMNIST dataset our implementation (FM3) shows a 1.35\% increase in prediction accuracy over the best-performing individual model it includes (i.e Resnet44) whereas the big/little configuration (FM3) shows only an 0.06\% increase in prediction accuracy compared to the single large model (i.e. RepVGG-A2).

%To further evaluate the impact of complementarity on the accuracy of CNN pairs, we conducted
The experiments using six pairs of CNNs with different complementarity values are illustrated in Figure \ref{fig:coverage-gain}. The green and orange segments of the stacked bars represent the accuracy of the individual CNNs. The purple segment indicates the accuracy of CNN pairs based on our proposed methodology. The numbers inside the stacked bar charts show the accuracy gain of the CNN pairs compared to the best-performing individual CNN.
The outcomes show that utilizing a pair of CNNs with a relatively higher complementarity factor leads to improved accuracy. This is achieved by covering a larger portion of the dataset, despite employing smaller and less complex models as we described in the subsection \ref{SubSec:ProposedMethodology_2Het}.

\subsubsection{Response Time Evaluation}
Additionally, our methodology improves response times, as shown in Figure \ref{fig:response}. 
% For the CIFAR-10 dataset, our configuration (C3) has 52.7\% lower mean response time, 44.6\% lower 95-th percentile tail latency and  43.7\% lower 99-th percentile tail latency compared to the big/little configuration (\textcolor{blue}{C4}). Similarly it has 56.9\% lower mean response time, 2.5\% lower 95-th percentile tail latency and  2.5\% lower 99-th percentile tail latency compared to the single large model (C1). 
% We have the same observations with the ImageNet dataset: our configuration (\textcolor{blue}{I5}) has 65.2\% lower mean response time, 49.5\% lower 95-th percentile tail latency and  49\% lower 99-th percentile tail latency compared to the big/little setup (\textcolor{blue}{I4}) and  60.3\% lower mean response time, 22.9\% lower 95-th percentile tail latency and  21.1\% lower 99-th percentile tail latency compared to the single large model (I1).
%major revision
For the CIFAR-10 dataset, our configuration (CI3) has 52.7\% lower mean response time, 44.6\% lower 95-th percentile tail latency and  43.7\% lower 99-th percentile tail latency compared to the big/little configuration (CI2). Similarly it has 56.9\% lower mean response time, 2.5\% lower 95-th percentile tail latency and 2.5\% lower 99-th percentile tail latency compared to the single large model (CI1). Compared to the large pruned model (CI6) it has 41.1\% lower mean response time but a 26.1\% higher 95-th percentile tail latency and 24.6\% higher 99-th percentile tail latency. blue{Compared to the small distilled model (CI7) it has 47.3\% lower mean response time but 17.1\% higher 95-th percentile tail latency and 16.4\% 99-th percentile tail latency.}

We have the same observations with the ImageNet dataset: our configuration (IM3) has 65.2\% lower mean response time, 49.5\% lower 95-th percentile tail latency and 49\% lower 99-th percentile tail latency compared to the big/little setup (IM2) and  60.3\% lower mean response time, 22.9\% lower 95-th percentile tail latency and  21.1\% lower 99-th percentile tail latency compared to the single large model (IM1). Compared to the large pruned model (IM6) it has 31.6\% lower mean response time but 33.8\% higher 95-th percentile tail latency and 34.5\% increase in 99-th percentile tail latency.

For the Intel dataset, our configuration (IN3) achieves a 27.4\% reduction in mean response time, a 17.6\% decrease in 95th percentile tail latency, and a 17.4\% decrease in 99th percentile tail latency compared to the big/little setup (IN2). When compared to the single large model (IN1), our implementation shows an 18.2\% reduction in mean response time, though it exhibits a 48.9\% increase in 95th percentile tail latency and a 45.0\% increase in 99th percentile tail latency. Relative to the large pruned model (IN6), it features a 1.7\% decrease in mean response time but has 77.9\% higher 95th percentile tail latency and 76.9\% higher 99th percentile tail latency. Compared to the small distilled model (IN7), it shows a 5.7\% increase in mean response time, alongside an 80.6\% increase in 95th percentile tail latency and a 76.3\% increase in 99th percentile tail latency.

Lastly, for the FashionMNIST dataset, our configuration (FM3) delivers an 18.8\% reduction in mean response time, a 40.7\% decrease in 95th percentile tail latency, and a 35.8\% decrease in 99th percentile tail latency compared to the big/little setup (FM2). Compared to the single large model (FM1), our implementation shows an 11.9\% reduction in mean response time, but with a 40.0\% increase in 95th percentile tail latency and a 35.8\% increase in 99th percentile tail latency. In comparison to the large pruned model (FM6), it has a 9.6\% increase in mean response time, with 69.4\% higher 95th percentile tail latency and 70.3\% higher 99th percentile tail latency. Relative to the small distilled model (FM7), it exhibits a 9.8\% increase in mean response time, a 73.6\% increase in 95th percentile tail latency, and a 40.5\% increase in 99th percentile tail latency.

% Using two CNNs increases response time and tail latency compared to a single CNN as we can see in Table \ref{table:singles}. This happens because the second CNN is invoked when the first CNN's prediction confidence is low, adding an additional response delay. This effect is seen in both our methodology and the big/little configuration. Nevertheless, despite the increased tail latency, our implementation achieves the lowest overall response times compared to other configurations, with the mean response time and tail latency being significantly lower for both datasets.

%Last but not least, we observe that the response times using the CIFAR-10 are lower compared to ImageNet due to the smaller image sizes and fewer classes in CIFAR-10, which reduce computational complexity and processing requirements.

Using two CNNs results in an increase in response time and tail latency compared to a single CNN, as shown in Table \ref{table:singles}. This occurs because the second CNN is triggered when the first CNN's prediction confidence is low, introducing an additional delay in response time. This effect is observed in both our methodology and the big/little configuration. Despite the increased tail latency, our implementation achieves a lower overall response time when compared to configurations with similar performance, such as the single large model and the big/little configuration. In comparison to the pruned model and the small distilled model, our configuration shows a higher tail latency due to the extra time needed to invoke the second CNN however it's performance is higher.

\begin{table*}[]
\centering
\begin{tabular}{@{}llcccc@{}}
\toprule
 &
  \multicolumn{1}{c}{\multirow{2}{*}{\textbf{CNN Pairs}}} &
  \multirow{2}{*}{\textbf{\begin{tabular}[c]{@{}c@{}}Individual\\ Accuracy\end{tabular}}} &
  \multicolumn{3}{c}{\textbf{Accuracy using our methodology}} \\
 &
  \multicolumn{1}{c}{} &
   &
  \begin{tabular}[c]{@{}c@{}}Default \\ pre-trained\\ weights\end{tabular} &
  \begin{tabular}[c]{@{}c@{}}Fine-tuned\\ Approach (a)\end{tabular} &
  \begin{tabular}[c]{@{}c@{}}Fine-tuned\\ Approach (b)\end{tabular} \\ \midrule
1 & \begin{tabular}[c]{@{}l@{}}MnasNet-1.3\\ DenseNet-121\end{tabular}      & \begin{tabular}[c]{@{}c@{}}69,17\%\\ 69.65\%\end{tabular} & 70.33\% & 69.33\% & 70.31\% \\
2 & \begin{tabular}[c]{@{}l@{}}MobileNet-V3-Large\\ GoogleNet\end{tabular}  & \begin{tabular}[c]{@{}c@{}}66.94\%\\ 65.27\%\end{tabular} & 69.12\% & 71.20\% & 70.78\% \\
3 & \begin{tabular}[c]{@{}l@{}}MnasNet-1.3\\ GoogleNet\end{tabular}         & \begin{tabular}[c]{@{}c@{}}65.57\%\\ 65.27\%\end{tabular} & 68.01\% & 66.95\% & 68.65\% \\
4 & \begin{tabular}[c]{@{}l@{}}RegNet-X-400MF\\ RegNet-Y-400MF\end{tabular} & \begin{tabular}[c]{@{}c@{}}66.84\%\\ 67.12\%\end{tabular} & 69.39\% & 69.38\% & 70.82\% \\
5 & \begin{tabular}[c]{@{}l@{}}RegNet-Y-400MF\\ GoogleNet\end{tabular}      & \begin{tabular}[c]{@{}c@{}}67.12\%\\ 65.27\%\end{tabular} & 69.03\% & 69.03\% & 69.74\% \\ \bottomrule
\end{tabular}
\caption {Enhancing complementarity evaluation}
\label{table:fine-tuned-evaluation}
\end{table*}

\subsubsection{Perceptual Hashing Ablation}
Figure \ref{fig:memory} presents the energy consumption comparisons when using the memory component. The comparison includes a single large model (CI1, IM1, IN1 and FM1) and a big/little architecture (CI2, IM2, IN2 and FM2) against our implementation without the memory component (CI3, IM3, IN3 and FM3), our implementation using the memory component with the Difference Hash (DHash) method (CI4, IM4, IN4 and FM4), and our implementation using the memory component with the Invariants from Complex Moments  method (CI5, IM5, IN5 and FM5). 
The configurations without a memory component (CI1 to CI3, IM1 to IM3, IN1 to IN3 and FM1 to FM3) exhibit a linear increase in energy consumption as the number of duplicated samples increases. In contrast, our implementations utilizing the memory component (CI4, CI5, IM4, IM5, IN4, IN5, FM4, and FM5) show almost no increase in energy consumption, with the slopes of the corresponding lines being very close to zero. This indicates that leveraging the memory component in an environment with identical, mirrored and rotated inputs significantly reduces energy consumption. Additionally, it is noteworthy that while the Invariants method is robust to image transformations, it demands more computational and energy resources compared to the DHash method.

\begin{figure}[ht!]
\centering
\includegraphics[width=0.5\textwidth]{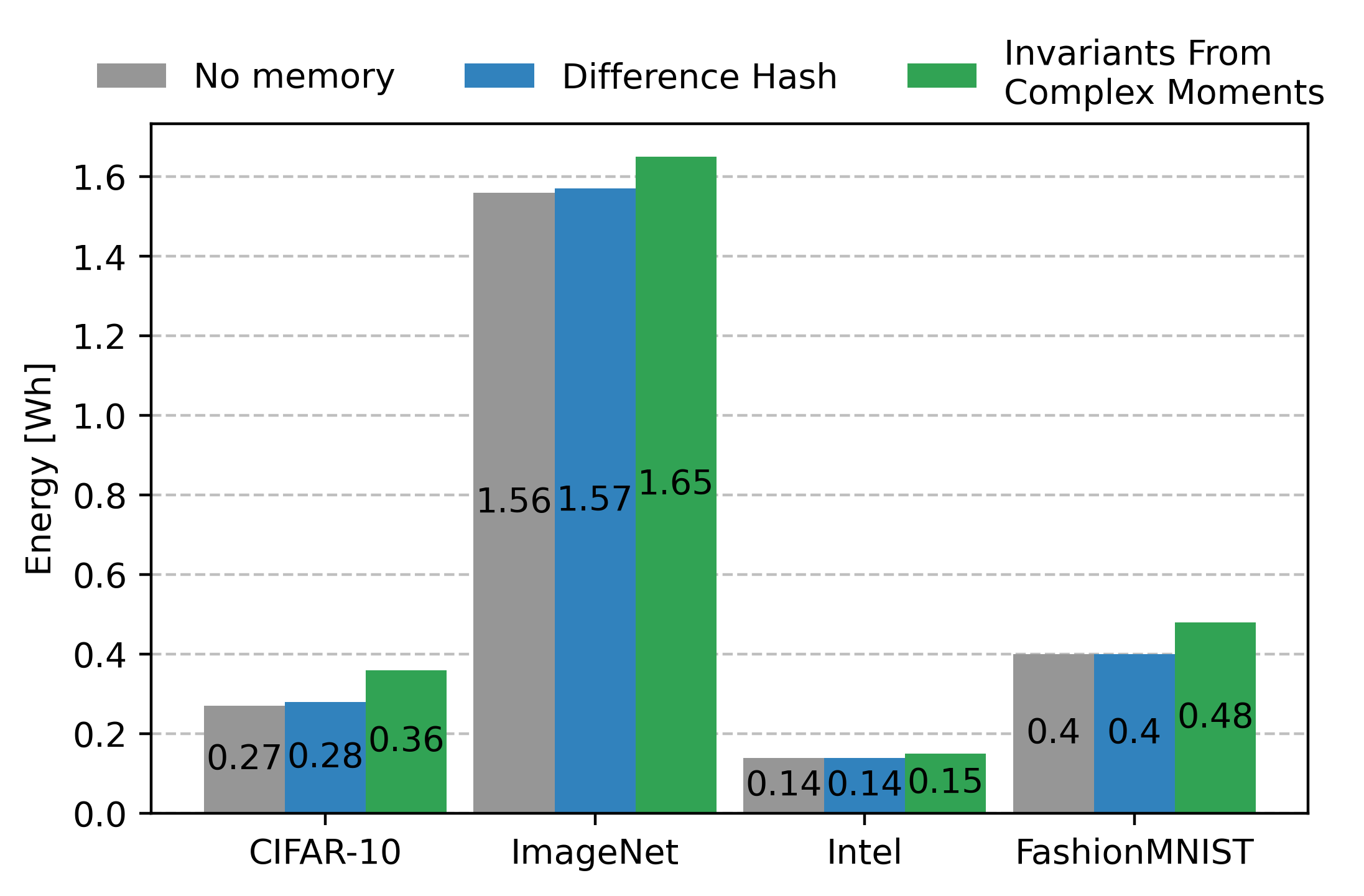}
\caption {Memory component computational overhead.}
\label{fig:memory_overhead}
\end{figure}

We also evaluated the computational overhead introduced by the memory component. To quantify this, we measured the energy consumption and current draw using the standard dataset without duplicated samples, allowing us to determine the additional energy required for the memory component's operation. As shown in Figure \ref{fig:memory_overhead}, for the CIFAR-10 dataset, the Difference Hash method led to a 1.9\% increase in energy consumption, while the Invariants From Complex Moments method resulted in a 30.9\% increase compared to not using the memory component. For the ImageNet dataset, energy consumption increased by 1.0\% with the Difference Hash method and by 5.8\% with the Invariants From Complex Moments. In the case of the Intel dataset, the Difference Hash method did not show any measurable increase in energy consumption, whereas the Invariants From Complex Moments led to a 7.1\% increase. Finally, for the FashionMNIST dataset, there was no measurable increase in energy consumption with the Difference Hash method, while the Invariants From Complex Moments increased energy consumption by 20\%.

\color{black}

% However, in the case where an already classified image, whose classification label exists in the hash table, has been rotated or mirrored and is re-inputted into the system, the DHash method will calculate a different fingerprint, and the image will be considered as not having been seen before. In contrast, the Invariants method will compute the exact same fingerprint, allowing the system to bypass the CNNs.

% \begin{figure}[ht!]
% \centering
% \begin{subfigure}[]{.5\textwidth}
%    \includegraphics[width=1\textwidth]{Figures/memory-cifar.png}
%    \caption{CIFAR10}
%    \label{fig:cifar10-memory} 
% \end{subfigure}

% \begin{subfigure}[]{.5\textwidth}
%    \includegraphics[width=1\textwidth]{Figures/memory-imagenet.png}
%    \caption{ImageNet}
%    \label{fig:imagenet-memory}
% \end{subfigure}
% \caption[Memory Component Comparison]{Energy consumption using the memory component for CIFAR10 (a) and ImageNet (b). Comparing our implementation (DC-2 \& DI-2) against the single large CNN (SC-1 \& SI-1) and the big-little (DC-1 \& DI-1)}
% \label{fig:energy measurements with memory}
% \end{figure}

%NEW Fine-Tuning
\subsubsection{Enhancing Complementarity Evaluation}

%Table \ref{table:fine-tuned-evaluation} summarizes the results of our methodology using the fine-tuned pairs with both approaches that increase complementarity as presented in subsection \ref{SubSec:IncreasingComplementarity}. 

Table \ref{table:fine-tuned-evaluation} summarizes the results of our methodology, using fine-tuned model pairs from both approaches outlined in subsection \ref{SubSec:IncreasingComplementarity} to enhance complementarity. The table presents five CNN pairs with their default weights, the individual accuracy of each CNN, the accuracy of each pair of CNNs, and the results after applying the first and second fine-tuning approaches. Some pairs exhibit marginally better performance with the first approach, while others perform slightly better with the second approach, with a performance gain of approximately 1\% in most cases. A notable result is observed with the second pair, where performance increased by 2.08\%, achieving an overall accuracy of 71.20\%, which surpassed even our main selected ImageNet configuration IM3 (see Table \ref{table:configurations}). Given that the highest-performing individual model in the pair achieved an accuracy of 66.94\%, this results in an overall accuracy improvement of 4.26\% using our methodology with fine-tuned models.

\begin{figure}[ht!]
\centering
\includegraphics[width=0.5\textwidth]{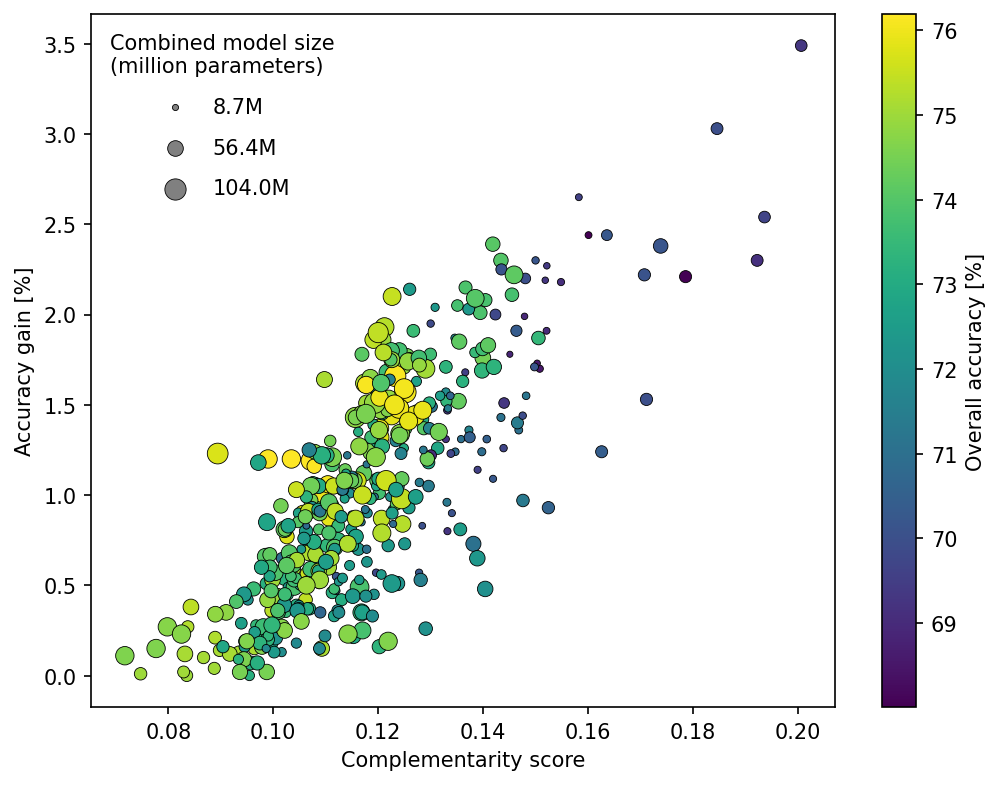}
\caption {Correlation between complementarity and accuracy gain}
\label{fig:complementarity-vs-accuracy-gain}
\end{figure}

%NEW Correlation
\subsubsection{Correlation Between Complementarity \& Increased Performance}

To better understand the correlation between complementarity scores and accuracy gains, we applied our methodology, as described in Section \ref{Sec:ProposedMethodology}, to all possible pair combinations of the ImageNet dataset models used in the complementarity matrix, to assess performance outcomes. The results are illustrated in Figure \ref{fig:complementarity-vs-accuracy-gain}. Each point on the graph represents a CNN pair, with the X-axis showing the complementarity score calculated using the complementarity formula (\ref{complementarity}), and the Y-axis showing the accuracy gain—defined as the additional accuracy achieved through our methodology compared to the best-performing model in the pair. %We observe a consistent trend where an increase in the complementarity score leads to a corresponding improvement in accuracy and model pairs having fewer parameters show notable accuracy gains. 

\begin{figure}[ht!]
\centering
\includegraphics[width=0.5\textwidth]{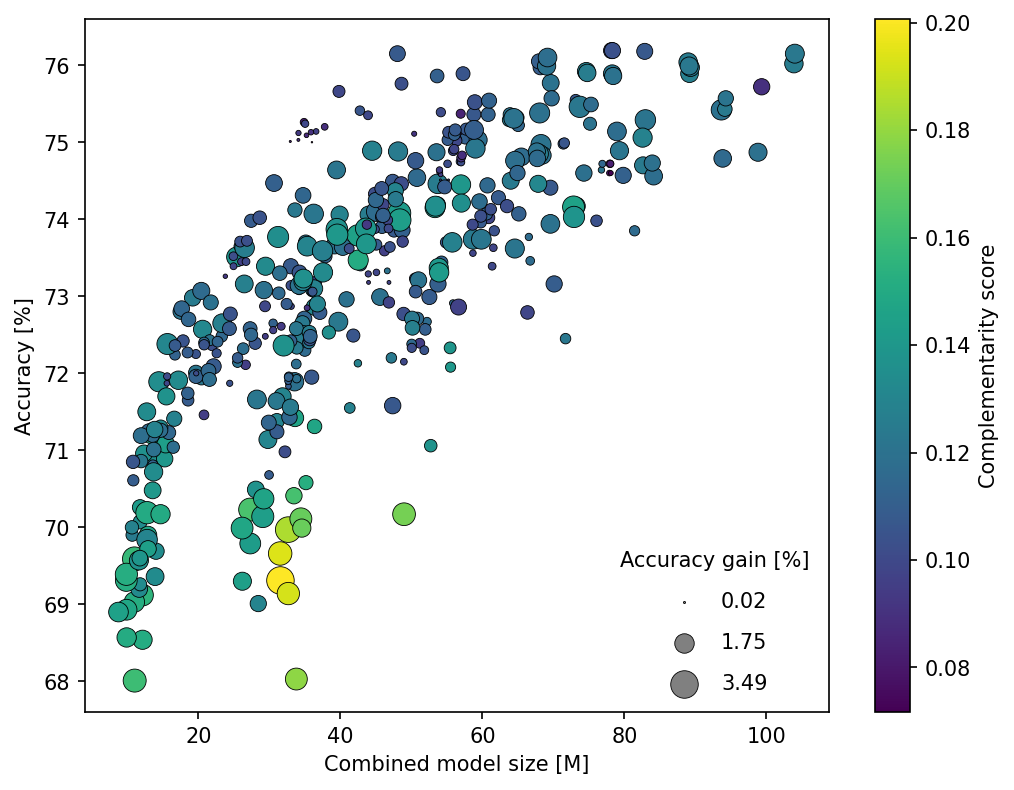}
\caption {Correlation between combined model size and overall accuracy}
\label{fig:model-size-accuracy}
\end{figure}

To provide further insight into the correlation, we added two additional dimensions of information: the size of each point reflects the combined number of parameters (M) of both models, and the color indicates the overall accuracy achieved by the pair. The graph reveals a positive linear correlation between complementarity scores and accuracy gains, suggesting that utilizing more complementary CNN pairs leads to improved accuracy. When examining overall performance, the largest models achieve the highest absolute accuracy, as expected. Despite this, their low complementarity scores result in minimal accuracy gains. This indicates tha, in addition to their higher energy demands, these large pairs of models are also less efficient, as a single large model would provide nearly the same level of performance.

To further explore the nature of complementarity, in Figure \ref{fig:model-size-accuracy}, we plotted the same results but with different axes. In this graph, the X-axis represents the combined model size in millions of parameters (M), and the Y-axis represents the overall accuracy achieved by the pair using our methodology. Additionally, the size of each point corresponds to the accuracy gain, while the color indicates the complementarity score. The graph reveals a logarithmic correlation, rather than a linear one, between model size and performance. This implies that although larger models tend to improve accuracy, the gains become progressively smaller as size increases, leading to a higher energy cost per unit of accuracy.%This suggests that while increasing model size generally improves accuracy, the returns diminish as size increases, leading to a higher energy cost per unit of accuracy.
Based on these insights, we can conclude that using a highly complementary pair of smaller CNN models results in increased accuracy with lower energy demand.

\subsection{Discussion} 
\label{SubSec:ExpEval_Discussion}

Our methodology contributes to advancing knowledge in the field of energy-efficient on-device AI applications by making research on a relatively unexplored approach: dynamically switching between small DNNs based on confidence scores. This strategy significantly enhances energy efficiency while maintaining predictive accuracy, offering a novel solution for resource-constrained smart environments.
Two innovative aspects of our approach have the potential to shape future research directions. First, we illustrate how an effective collaboration between complementary DNNs can improve prediction accuracy by leveraging the strengths of each model to compensate for the weaknesses of the other. This work demonstrates the potential for DNNs to collaborate on the same task, setting the stage for further exploration of cooperative model architectures. Second, we highlight the importance of integrating a memory component that retains previous decisions, allowing the system to bypass redundant inference processes for repeated inputs. This synergy between inference and memory opens new opportunities for optimizing both energy efficiency and predictive performance in AI-driven edge computing environments.

In terms of effectiveness, unlike traditional compression techniques or the big-little architecture, the proposed dual complementary CNN methodology leverages two smaller models that compensate for each other's weaknesses. This design achieves accuracy comparable to larger, more complex models while significantly reducing energy consumption. By dynamically selecting predictions based on confidence scores, the method ensures that only the most accurate predictions are used, enhancing overall inference reliability. This approach effectively maintains high accuracy without the computational overhead typically associated with large models.

In terms of efficiency, the methodology is specifically designed to reduce energy consumption and response time, a critical concern for edge devices with limited power and computational resources. The dual-CNN setup selectively activates the second CNN based on prediction confidence, ensuring it is only invoked when necessary. Additionally, the integration of a memory component, which stores and recalls previous classifications, further reduces energy consumption by bypassing repeated inference on identical or similar inputs.%Experiments on the Jetson Nano demonstrated energy savings of up to 85.8\% on CIFAR-10 and 80.9\% on ImageNet, marking a significant improvement over conventional CNN deployment strategies.

In terms of applicability, the methodology is broadly suitable for resource-constrained devices, such as those in IoT, smart homes, or edge computing environments. Unlike hardware-specific optimizations, such as co-design approaches requiring custom hardware or intensive model compression techniques (e.g., pruning, quantization), this dual-CNN method is hardware-agnostic and can be deployed on a wide range of devices without substantial modifications. The use of perceptual hashing in the memory component also makes it ideal for dynamic environments where inputs may frequently repeat or vary slightly, providing practical benefits in real-world scenarios.

%\textcolor{blue}{The first step of our methodology involves applying the complementarity formula to a list of pretrained models, with the objective of identifying a pair that covers the majority of dataset predictions while keeping model size minimal. Complementarity is a useful general indicator of which model pairs are likely to enhance performance. As shown in Figure \ref{fig:complementarity-vs-accuracy-gain}, there is a positive correlation between higher complementarity and performance gain. Although reduced overlap can indicate models that generalize less, our experiments show that does not necessarily lead to poorer performance, but the opposite. Additionally the goal of this paper is not to modify models explicitly to decrease overlap, but using the complementarity formula, provide a general direction for pair selection.} 

Our proposed methodology focuses on multi-class classification, but not multi-label classification, specifically using visual data. At present, it does not support data samples that can be assigned multiple labels or categories simultaneously. Our design has the limitations of being tailored and evaluated for visual data, opening opportunities for further research to explore other data modalities such as audio, biometric, geo-location, and sensor time series data in smart environments. Additionally, we acknowledge the limitation that our methodology operates on two small prediction models, and if such models are unavailable, a transfer learning process will be necessary. In cases where the transfer learning models have low accuracy, our experimental results demonstrate that our methodology significantly improves their performance.

To ensure the robustness and validity of our experimental results, we conducted experiments using four different datasets. We carefully managed data splitting by employing random shuffling, stratification, and preventing data leakage. The evaluation was performed on unseen data to provide an unbiased performance measure. Furthermore, we compared the performance of our proposed model against well-established baseline models, including pruning, knowledge distillation, and the big/little model \cite{park_biglittle_2015}. We also ensured the reproducibility of our experiments by testing with a variety of architectural pairs, thereby confirming the consistency of the improvements observed across different model synergies.

\section{Conclusion and Future Work}
In this study, we propose a methodology for reducing the energy requirements of on-device CNNs inference through the utilization of two complementary CNNs integrated with a memory component. Each CNN addresses the weaknesses of the other, while the memory component retains previous classifications, thereby bypassing the need for repeated CNN invocations. Our implementation demonstrates up to 85.8\% reduction in energy consumption compared to a single large CNN for inferences with multiple identical inputs on the CIFAR-10 dataset up to 80.9\% energy reduction on the ImageNet dataset, up to 76.0\% energy reduction on the Intel dataset and up to 77.5\% energy reduction on the FashionMNIST dataset with negligible loss of accuracy.

Our future work involves adapting the concept of complementarity to various data modalities and applications in smart environments.
Further research directions include examining complementarity based not only on the number of predictions made by the models in a validation dataset but also by focusing solely on confidence scores. Additionally, we aim to develop a new complementarity formula that considers the intrinsic structural characteristics of the CNNs. Lastly, since the concept of complementarity is a significant contribution of our work, We aim to explore its application for enhancing accuracy, rather than solely focusing on reducing energy consumption.

%% If you have bibdatabase file and want bibtex to generate the
%% bibitems, please use
%%

\section*{Acknowledgment}
This work was funded by the European Union’s Horizon Europe research and innovation program under grant agreement No. 101120237 (ELIAS)

 \bibliographystyle{elsarticle-num} 
 \bibliography{References}

%% else use the following coding to input the bibitems directly in the
%% TeX file.

% \begin{thebibliography}{00}

% %% \bibitem{label}
% %% Text of bibliographic item

% \bibitem{}

% \end{thebibliography}
\end{document}